\documentclass{article}

\usepackage{arxiv}

\usepackage[utf8]{inputenc} 
\usepackage[T1]{fontenc}    
\usepackage{hyperref}       
\usepackage{url}            
\usepackage{booktabs}       
\usepackage{amsfonts}       
\usepackage{nicefrac}       
\usepackage{microtype}      
\usepackage{lipsum}
\usepackage{graphicx}
\usepackage{amsmath,amsfonts}
\usepackage{algorithmic}
\usepackage{algorithm}
\usepackage{array}
\usepackage[caption=false,font=normalsize,labelfont=sf,textfont=sf]{subfig}
\usepackage{textcomp}
\usepackage{stfloats}
\usepackage{url}
\usepackage{verbatim}
\usepackage{graphicx}
\usepackage{cite}
\usepackage{xcolor}
\usepackage{amsmath}
\usepackage{makecell}
\usepackage{multicol}
\usepackage{multirow}
\usepackage{booktabs}
\usepackage{siunitx}
\usepackage{gensymb}
\usepackage{soul, color, xcolor}
\usepackage[utf8]{inputenc}
\usepackage[english]{babel}
\usepackage{threeparttable}
\usepackage{graphicx} 
\usepackage{amsmath}
\usepackage{amssymb}
\usepackage{tablefootnote}
\usepackage{stfloats}
\usepackage{threeparttable}
\usepackage{color,soul}
\graphicspath{ {./images/} }

\title{Development and Characteristics of a Highly Biomimetic Robotic Shoulder Through Bionics-Inspired Optimization}

\author{
Haosen Yang \\
  The Department of Mechanical, Aerospace and Civil Engineering\\
  University of Manchester\\
  Manchester, M13 9PL, UK \\
  \texttt{haosen.yang@postgrad.manchester.ac.uk} \\
   \And
 Guowu Wei \\
  School of Science, Engineering and Environment\\
  University of Salford\\
  Salford, M5 4WT, UK \\
  \texttt{g.wei@salford.ac.uk} \\
  \And
Lei Ren \\
  The Key Laboratory of Bionic Engineering, Ministry of Education\\
  Jilin University\\
  Changchun 130025, China \\
  \texttt{lren@jlu.edu.cn} \\
}

\begin{document}
\maketitle
\begin{abstract}
This paper provides a comprehensive analysis of the existing landscape of conventional and highly biomimetic robotic arms, highlighting a prevalent trade-off between size, range of motion, and load capacity in current highly biomimetic designs. To overcome these limitations, this paper undertakes an in-depth exploration of the human shoulder, focusing on the identification of mechanical intelligence within the biological glenohumeral joint such as the incomplete ball-and-socket structure, coupling stability of humeroradial and glenohumeral joints, the self-locking mechanism of the glenohumeral joint. These intelligent features potentially enhance both the stability and mobility of robotic joints, all the while preserving their compactness. To validate these potential benefits, this paper introduces a novel, highly biomimetic robotic glenohumeral joint that meticulously replicates human musculoskeletal structures, from bones and ligaments to cartilage, muscles, and tendons. This novel design incorporates the mechanical intelligence found in the biological joint. Through rigorous simulations and empirical testing, this paper demonstrates that these principles significantly enhance the flexibility and load capacity of the robot's glenohumeral joint. Furthermore, extensive manipulation experiments confirm the robustness and viability of the completed highly biomimetic robotic arm. Remarkably, the presented robotic arm realised 46.25\% glenohumeral flexion/extension, 105.43\% adduction/abduction and 99.23\% rotation, and can achieve a payload of 4 kg, and open the door which requires a torque of over 1.5 Nm to twist the handle. Not only does this paper validate the innovative mechanical intelligence identified in the deconstruction of the human shoulder joint, but it also contributes a pioneering design for a new, highly biomimetic robotic arm, significantly pushing the boundaries of current robotic technology.
\end{abstract}


\section{Introduction}

Biomimetic robots, mirroring human kinetics, show significant potential in the Human-Robot Interaction (HRI) field. Their anthropomorphic design fosters adaptability to human-centric environments, enhancing usability and societal acceptance. While a variety of robots have emerged over the past {30 years}, challenges in biomimetic design persist, particularly in crafting humanoid glenohumeral joints. This study addresses this issue and proposes an innovative biomimicry-based solution.

Existing robotic arm designs frequently employ multiple rotary joints to simulate human shoulder movement. Although this approach simplifies the design and provides extensive motion range, it also demands substantial space, thereby limiting its integration into compact systems. On the other hand, humanoid robot arms utilize a single joint with multiple rotational degrees of freedom, offering a compact design but imposing restrictions on joint torque or size. These designs often rely on rigid materials for precision, potentially compromising safety in HRI scenarios.

In contrast, the human shoulder joint exemplifies a compact, highly mobile structure that strikes a balance between mobility and stability, showcasing substantial load-bearing capacity. It can generate high torque and possesses damping and elastic characteristics, enhancing safety and resilience. The capacity of the biological shoulder joint to dislocate and recover under extreme forces could potentially augment safety measures for robotic systems during close-range human interactions. {Nonetheless, within the normal load range, the joint maintains stability.}

Existing biomimetic research has yielded designs that mimic human structures, often leveraging a tendon-driven approach. However, these designs {(no rigid shafts and hinge joints are used)} typically offer an incomplete representation of human anatomy, particularly concerning soft tissues, leading to potential structural stability issues. Specifically, the mobility-stability trade-off in certain biomimetic robotic arms results in a limited range of joint motion. {A refined representation of soft tissues can enhance both load capacity and safety in robotic designs. Furthermore, such incorporation reduces oscillations during mechanical movements, as these robotic joints inherently demonstrate rotational resistance due to the soft tissues.} Greater attention must be devoted to these trade-offs and the intricate role of soft tissues in future design to enhance biomimetic potential.

Currently, highly biomimetic robotic arms primarily replicate the appearance and movement characteristics of human joints, often neglecting the complex benefits of human structure, such as the function of ligaments, tendons, and cartilage. This leads to a considerable gap between robotic and human anatomy. This research embarks on an exploration of the mechanical intelligence inherent in human joints, through a comprehensive deconstruction study, with the goal of identifying areas of opportunity to enhance robotic performance. The insights derived from the deconstruction study will be applied to the proposed robot design, with the aim of more accurately emulating the human joint. The development of such highly biomimetic robots has the potential to revolutionize current robotic designs, validate human tissue function, and further corroborate the findings of the deconstruction study. This research constitutes the first attempt to elucidate the functionality and superiority of various human anatomical structures via physical robotic prototypes, thereby bridging the divide between anatomical understanding and practical applications.

\section{Related work}

\subsection{Existing highly biomimetic shoulder designs}

Common shoulder joint designs for robotic arms often use multiple rotary joints in series to achieve three-degree rotational freedom\cite{walker1994modular,borst2009rollin}{(The opposite is the use of a universal joint)}, a strategy seen in industrial robots. This modular approach offers several advantages, such as simplified design, streamlined manufacturing and maintenance, wide motion range, and potentially infinite torque with the use of high-performance motors without limiting the limb size. However, a significant downside to this strategy is the substantial space it requires due to the need for the distance between individual joints for geared motor installation. This results in a larger overall shoulder size, which complicates the creation of compact, space-efficient robotic systems and may lead to bulky robotic arms that are impractical in limited spaces or for close human-robot interactions.

Humanoid bionic robotic arms frequently employ a single joint to achieve multiple rotational degrees of freedom within the shoulder joint\cite{grebenstein2011dlr,englsberger2014overview,paik2012development}. {A typical design incorporates a spherical joint providing two rotational degrees of freedom. This is supplemented by a bifurcated upper arm, where one segment is anchored to the ball joint while the other permits relative lateral/internal rotation.} This configuration retains the advantages of multiple degrees of freedom in series while simultaneously achieving a compact design. Nevertheless, this design approach introduces its unique set of challenges. The motor often necessitates positioning near or within the spherical joint itself, inevitably leading to a compromise in either the joint torque or joint size. This requisite often results in a reduction of joint torque, given the limited space available for motor installation. Alternatively, to maintain torque, the joint size may have to be increased to accommodate a larger motor, leading to a bulkier design. Moreover, such an arrangement often diverges substantially from the natural appearance of the biological shoulder. {The mechanical structure and overt components can impart a distinctly robotic appearance, which may be unsuitable for applications necessitating a humanoid look, such as displaying humanoid robots, prosthetics, and certain film props.}

This prevalent design approach predominantly relies on rigid materials to achieve superior stiffness and precision, crucial for the accurate and reliable operation of robotic arms. However, this focus on rigidity and precision often comes at the expense of safety, particularly in the context of HRI.

The human shoulder joint embodies a number of exceptional advantages that can be leveraged for the development of advanced robotic and automated systems. Compactness and Mobility: The human shoulder joint, specifically the glenohumeral joint, is compact yet highly mobile. Capable of achieving three stable degrees of freedom and an extensive range of motion, this joint stands as the most mobile in the human upper limb. This compactness and mobility in one structure allow for dynamic movement capabilities in a confined space. Stability: The shoulder joint maintains a balance between mobility and stability. While being extremely flexible, it is also resilient, resisting easy dislocation {(can bear considerable loads)}. This enables the handling of strenuous tasks without interruption, thereby demonstrating significant load capacity. In most mechanisms, mobility and stability often exist as conflicting performance indicators; however, the glenohumeral joint exemplifies a unique balance between these two properties. High Torque Output: The human shoulder joint can generate substantial torque, a feature that would be beneficial for robotic systems involved in heavy-duty tasks. Safety and Compliance: Unlike traditional rigid joints, the human shoulder joint exhibits damping and elastic characteristics. This compliance allows for a degree of flexibility and resilience, providing safety against sudden external forces. {Self-Recovery Mechanism: Biological joints exhibit controlled dislocation under extreme forces, serving as an injury mitigation strategy. Translating this to robotics, particularly in designs for close human interaction, could enhance safety. Implementing joints with intentional dislocation capabilities, followed by repositioning similar to orthopaedic realignments, may reduce risks in unintended collisions or forceful interactions, example includes} \cite{seo2019human}

Exploring these features for integration into a robotic arm system could significantly enhance safety, particularly in human-robot interaction environments, and provide superior performance in a compact form. The potential of these inherent human joint properties to advance robotic and automated systems deserves more in-depth exploration.

The study of biomimetics has captivated many researchers, leading to designs that emulate human biological structures \cite{hosoda2012anthropomorphic,seo2019human,diamond2012anthropomimetic,mizuuchi2006development,sodeyama2008designs,mouthuy2022humanoid,trendel2018cardsflow}. Numerous designs adopt a tendon-driven approach, akin to that of a biological arm, utilizing the inherent physical properties of tendons to imitate the compliance and dynamic characteristics of the musculoskeletal system. This approach often culminates in designs that bear greater resemblance to a biological arm.

Existing designs of robotic arms, despite successfully emulating fundamental human shoulder functions, often present only a partial replication of human anatomy due to the oversimplification of biological principles, consequently compromising performance. Modern musculoskeletal robotic arms typically model the glenohumeral joint as a spherical joint, prioritizing stability by fully enclosing the ball within the socket. However, this simplification restricts the joint's range of motion. In robotic models such as Kengoro\cite{asano2017design} and Kenshiro\cite{asano2019musculoskeletal}, stability of the glenohumeral joint is prioritized at the cost of certain functional aspects. For instance, Kengoro achieves only 50\% of the biological shoulder's range of motion for shoulder flexion/extension, while Kenshiro provides a mere 10\% of the range of motion for shoulder rotation. Conversely, the human glenohumeral joint adopts a level of instability to extend the motion range. The humeral head is significantly larger than the glenoid and not entirely enclosed by it, thus allowing greater joint mobility.

The lack of soft tissue representation in these models contributes to a significant deviation from the biological human form. An elaborate inclusion of soft tissue can enhance load capacity, compliance, and flexibility at joint extremities. Soft tissues also offer a recovery mechanism following dislocation due to extreme external forces, augmenting safety in HRI. These tissues serve as critical elements in damping oscillations during mechanical movements, suggesting their complex role that requires careful consideration to fully harness the potential of biomimetic design.

Conventional robotic arms utilize bearings and shafts to resist multidirectional impacts and mechanical structures to limit motion, depending on material strength for precision and load capability. Despite substantial advancements in mirroring the functionality and morphology of human arms, contemporary biomimetic robots still rely predominantly on traditional mechanical connections. These designs may lack compliance and exhibit reduced resilience to fluctuating loads. Current musculoskeletal robotic arms may also demonstrate restricted load-carrying capacity and vulnerability to joint dislocation. This joint instability during movement is evident in the associated presentation video\cite{potkonjak2011puller}. In contrast, the human body integrates tension and compression structures, with soft tissues like cartilage, ligaments, and tendons primarily limiting motion. These tissues function as a low-pass filter, absorbing forces without causing dislocation or damage. The succeeding subsection will examine the biological structures that contribute to impact mobility and improved load-bearing capacity.

\subsection{Anatomy study - Mechanical intelligence}

The human shoulder, an intricate system enabling complex arm motion and expansive reach, comprises three bones (the clavicle, scapula, and humerus) and four joints. The glenohumeral joint, integral to shoulder mobility, offers three degrees of freedom and approximately 2/3 of the shoulder's motion range, the scapulothoracic joint provides the residual 1/3\cite{aliaj2022kinematic}. Both sternoclavicular and acromioclavicular joints\cite{giovannetti2023evaluation} contribute to shoulder stability, forming a triangular linkage among the clavicle, scapula, and torso. This paper primarily focuses on the glenohumeral joint.


\begin{figure}[htb]
\centerline{\includegraphics[width=0.65\textwidth]{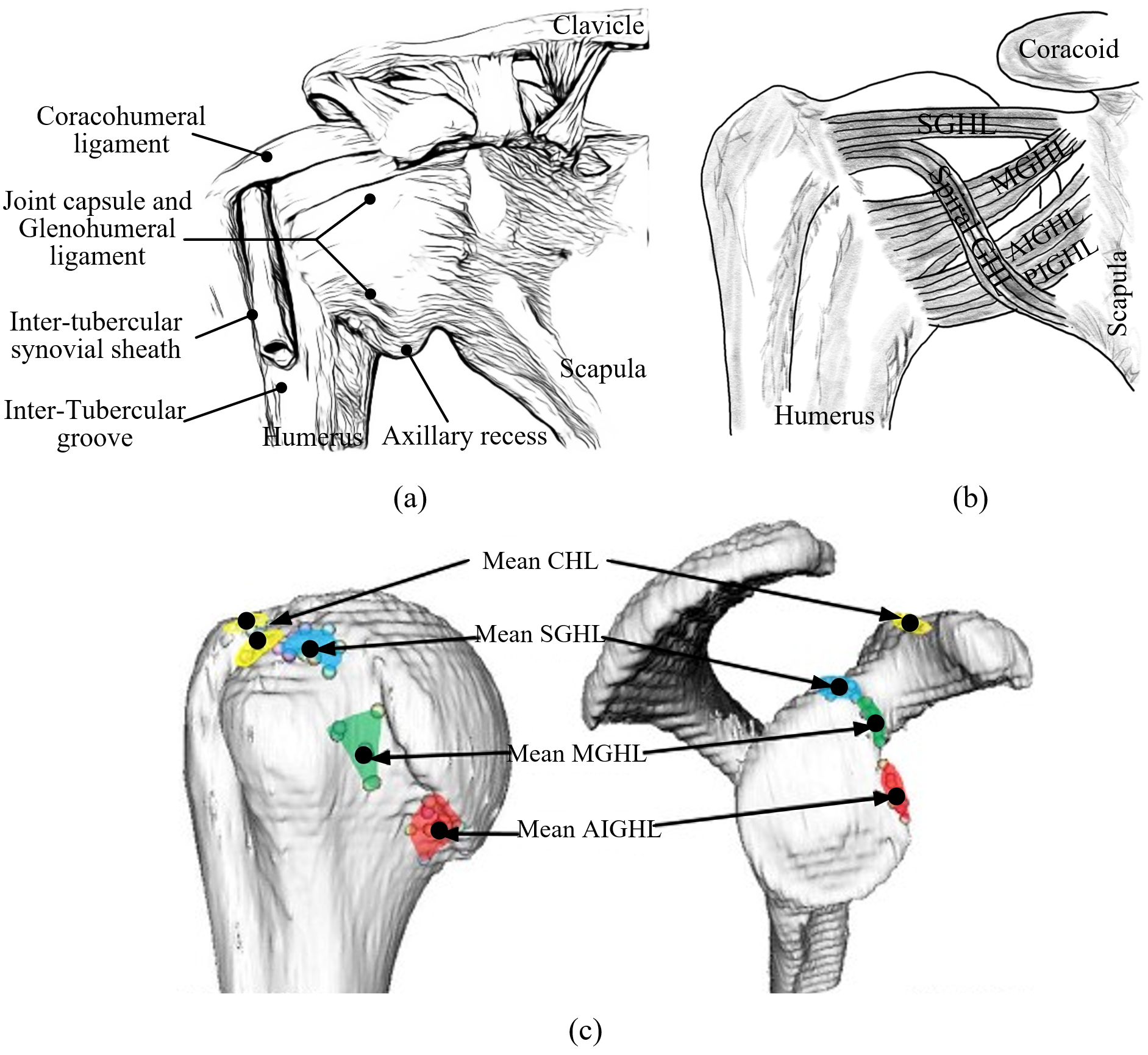}}
\caption{(a) The capsule and ligament of the glenohumeral joint (b) glenohumeral ligament\cite{felstead2017biomechanics} (c) Mean insertion of glenohumeral ligament and coracohumeral ligament\cite{yang2010vivo}.}
\label{fig2.13}
\end{figure}

The glenohumeral joint is the most mobile joint in the upper limb\cite{rugg2018surgical}. It has remarkable mobility, primarily facilitated by seven ligaments, which only restrict the joint's motion within certain limits. The coracohumeral ligament (CHL) and glenohumeral ligament (GHL)(Fig.\ref{fig2.13}(a), (b)) play key roles in limiting joint motion. CHL (anterior and posterior) restricts anterior-inferior translation during joint rotation\cite{falworth2018biomechanics}, whereas GHL (superior-SGHL, middle-MGHL, and inferior-IGHL) provides stability during arm adduction and abduction. The GHL's inferior band (IGHL) with its anterior (AIGHL) and posterior subcomponents (PIGHL) contribute to stability during 90\degree of flexion and internal rotation when the shoulder is abducted\cite{frank2015rotator, felstead2017biomechanics}. The locations at which the ligaments are inserted are shown in Fig.\ref{fig2.13}(c). The intra-articular nature of the glenohumeral joint is attributed to its joint capsule\cite{itoi1996biomechanical}. 

Within this motion range, the ligaments are taut, and the joint is ‘unstable'. The joint's ball and socket structure with a small socket enables a large range of motion for each of the three degrees of freedom. The high-performance capabilities of the glenohumeral joint are underscored by the following distinctive features. 

\noindent\subsubsection{\textbf{The humeral head's articular surface substantially surpasses the scapula's}}

\begin{figure}[htb]
\centerline{\includegraphics[width=0.7\textwidth]{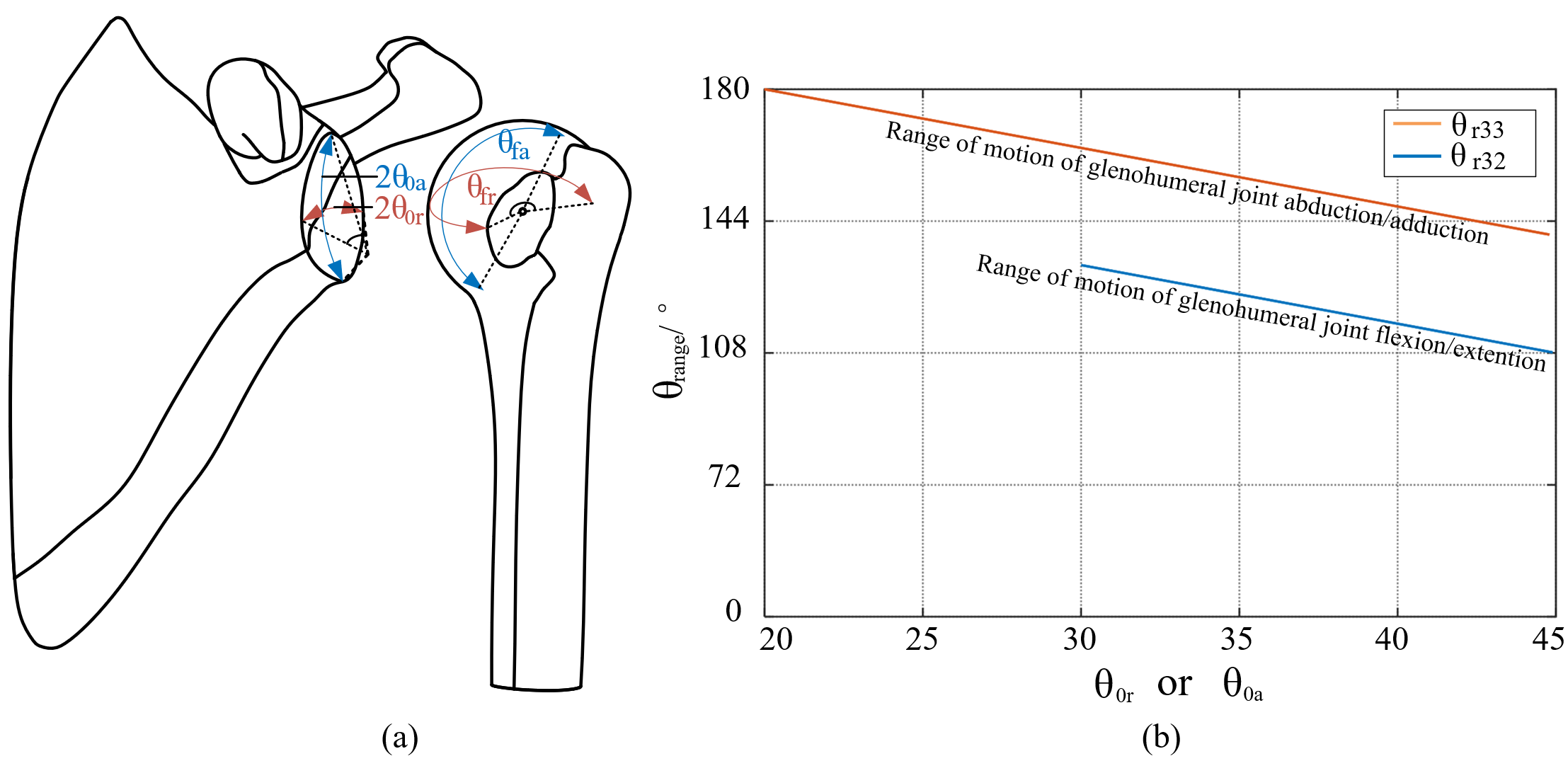}}
\caption{(a) The size of humeral head and glenoid. (b) The relation between $\theta_{0r}$, $\theta_{0a}$ and $\theta_{r33}$, $\theta_{r32}$.}
\label{fig6.20}
\end{figure}

Research data \cite{churchill2001glenoid} reveals the humeral head to be larger, with the glenoid's width and height being 62.5\% and 73.5\% of the humeral head's, respectively. \cite{mcpherson1997anthropometric} confirmed this founding, suggesting a greater coronal constraint that limits superior-inferior translation while facilitating sagittal movement. According to Fig. \ref{fig6.20}(a), the range of motion of glenohumeral joint rotation is $\theta_{r33}=\theta_{fr}-2\theta_{0r}$ ($\theta_{fr}$ and $\theta_{0r}$ are described in figure). The range of motion of joint abduction/adduction is $\theta_{r32}=\theta_{fa}-\theta_{0a}$ ($\theta_{fa}$ and $\theta_{0a}$ are described in figure). Figure \ref{fig6.20}(b) illustrates a decrease in $\theta_{r33}$ and $\theta_{r32}$ as $\theta_{0r}$ and $\theta_{0a}$ increase, providing evidence that increases the articular surface of the humeral head while reducing that of the scapula promotes enhanced joint mobility. The glenoid labrum addresses this mismatch by expanding the dimensions of the glenoid cavity, thus augmenting mobility while still sacrificing stability. In the biological glenohumeral joint, it enables a wide motion range as evidenced in Table. \ref{tab2.3}, and sufficient load capability is also achieved. 

\begin{table}[htb]
\caption{Range of motion for biological glenohumeral joint (data from \cite{asano2019musculoskeletal,nordin2001basic}).}
\footnotesize
\begin{center}
\begin{tabular}{l c}
\toprule
\makecell[c]{Motion group} & \makecell[c]{Range of motion} \\
\midrule
Glenohumeral Abduction (-)/Adduction (+) & 0-120\degree \\
Glenohumeral Flexion (-)/Extension (+) & -110\degree-60\degree \\
Glenohumeral Internal (-)/External (+) rotation & -97\degree-34\degree \\
\bottomrule
\end{tabular}
\label{tab2.3}
\end{center}
\end{table}

The load capacity in the biological glenohumeral joint is upheld through passive and active elements as the ligaments are only kept tightened when maximum joint motion is reached. Passive elements include humeral-glenoid conformity, the glenoid labrum, the peripheral thickening and loading-induced deformability of the glenoid cartilage, and negative intra-joint pressure, which generates a vacuum effect, securing the humeral head within the glenoid cavity. Active elements involve muscle forces pressing the humeral head into the glenoid fossa, mainly provided by the rotator cuff group. The subscapularis muscle, which creates a self-locking mechanism when the arm is adducted. Further, Multiple tendons crossing the elbow and glenohumeral joint provide additional stability. The biceps muscle's long head, attached to the glenoid labrum, adds a compressive force during biceps movement, aiding in joint stabilization. These mechanisms jointly sustain shoulder joint stability.

\subsubsection{\textbf{Negative intra-articular pressure}}
\begin{figure}[htb]
\centerline{\includegraphics[width=0.5\textwidth]{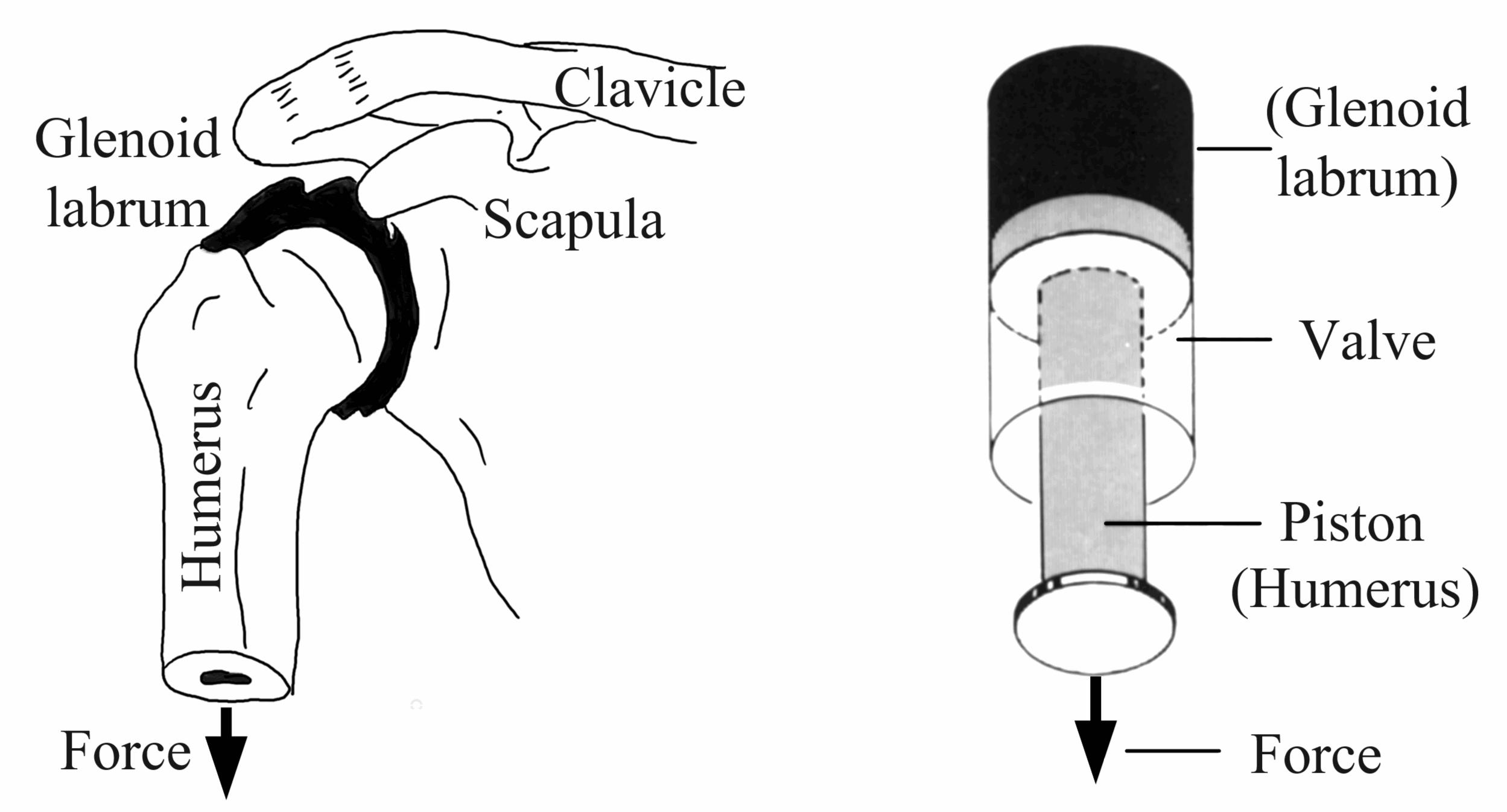}}
\caption{Cylindrical valve filled with fluid and the glenohumeral joint\cite{manson2000forearm}.}
\label{fig2.69}
\end{figure}

Researchers have discovered the crucial mechanical function of the glenoid labrum, which is to function as an anti-shear bumper\cite{kumar1985role, resnik1984intra, hashimoto1995dynamic}. When the humeral head is pressed into the glenoid fossa in the presence of an intact labrum, air or fluid is squeezed into the joint capsule, creating a negative pressure within the glenoid fossa. This creates a situation similar to a cylindrical valve filled with fluid (as illustrated in Fig.\ref{fig2.69}), which effectively stabilizes the glenohumeral joint. Habermeyer et al.\cite{habermeyer1992intra} examined the influence of atmospheric pressure on glenohumeral joint stabilization, demonstrating stability forces ranging from 68-225 N exerted by external atmospheric pressure on a cadaveric shoulder. This result is also supported by the study conducted by Nobuyuki Yamamoto et al.\cite{yamamoto2006effect}. Both researchers pointed out that the dynamic stability provided by muscular balance might also be influenced by the absence of negative intra-articular pressure. Combined with the results in the concavity-compression force, the difference with and without labrum may indicate the negative intra-articular pressure contributed up to 10\% of the stability ratio.

\begin{figure}[htb]
\centerline{\includegraphics[width=0.7\textwidth]{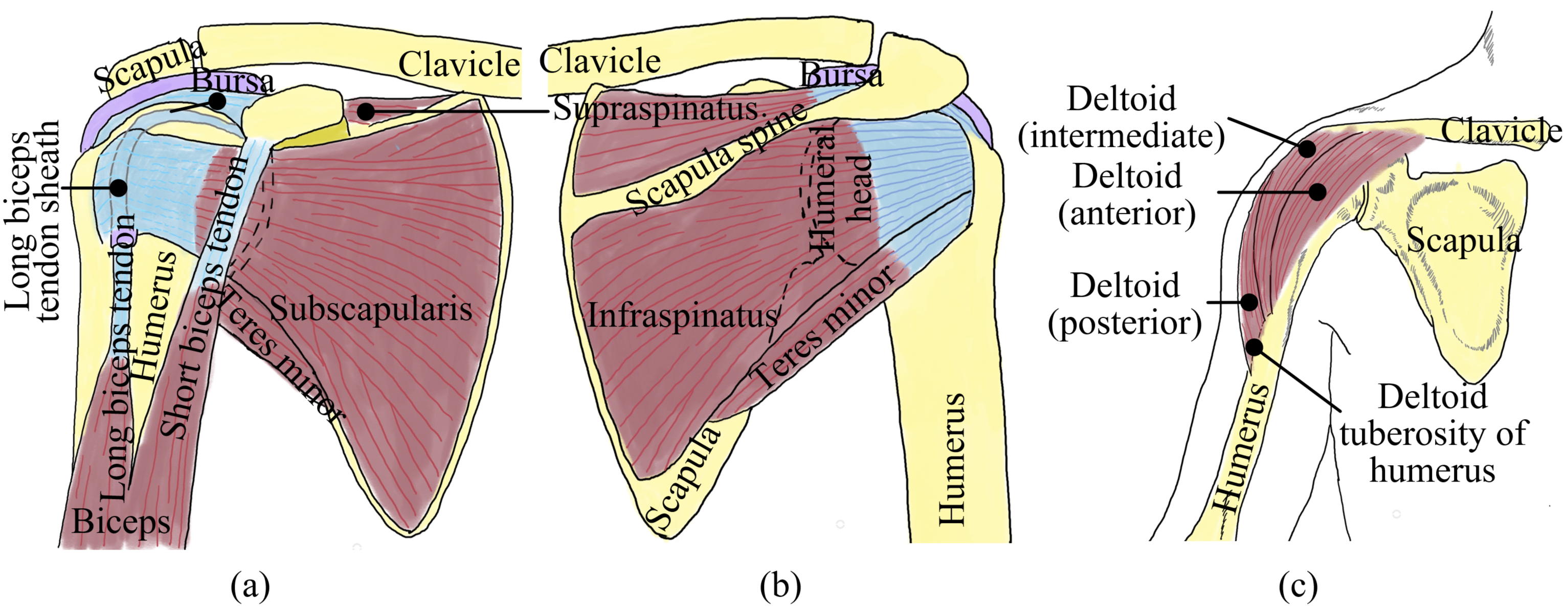}}
\caption{Muscle groups and tendons in the glenohumeral joint. Tendons are marked in blue, muscles are marked in red, bones are marked in yellow, the bursa is marked in purple (a) Rotator cuff (supraspinatus, subscapularis, teres minor and infraspinatus) and biceps muscle, front view. (b) Rotator cuff and biceps muscle, back view. (c) The deltoid muscle.}
\label{fig2.14}
\end{figure}

\subsubsection{\textbf{The rotator cuff muscle stabilize the joint}}
The dynamic stability of the glenohumeral joint during active movement relies heavily on surrounding tendons, with three main muscle groups contributing significantly: the rotator cuff muscles, the biceps tendon, and the deltoid muscle (Fig.\ref{fig2.14}). The rotator cuff, comprising supraspinatus, subscapularis, teres minor, and infraspinatus, dynamically stabilizes the shoulder joint by compressing the humeral head into the glenoid fossa during midrange shoulder movement, with ligaments playing a more pivotal role at motion extremes. With comparable cross-sectional depths for posterior and anterior rotators, an equal force couple prevents translation across the joint. The subscapularis muscle creates a self-locking mechanism when the arm is adducted. Under the condition of constant muscle length, an increase in the downward force applied to the humerus may enhance the stability of the glenohumeral joint. This concept will be subjected to a theoretical exploration in a subsequent section. 

\subsubsection{\textbf{Tendons across multiple joints}}

The long head of the biceps tendon (Fig.\ref{fig2.14}(a)), uniquely originating intra-articularly, traverses three joints: the glenohumeral, humeroulnar, and proximal radioulnar joints. Its loading during elbow flexion and forearm rotation provides a compression force to the glenohumeral joint, thus contributing to stability. Alexander et al.\cite{alexander2013role} examined this tendon's role in glenohumeral stability, comparing the anterior humeral head translation in an intact and vented capsule, under both loaded and unloaded conditions of the long head of the biceps tendon. Their findings suggest a considerable impact on the joint's overall stability by the loaded tendon, reducing anterior and inferior translations by 42.6\% and 73.3\%, respectively. Analogously, the long head of the triceps, which also crosses the elbow joint, may assist glenohumeral stabilization during elbow extension.

\section{Mechanical design and prototype}

This section elaborates on the design specifics of the proposed highly biomimetic glenohumeral joint. The overarching goal of this design is to emulate the human joint structure as precisely as possible, incorporating the previously detailed mechanical intelligence. This is intended to validate the functionality and spotlight the advantages of this mechanical intelligence, thereby addressing the challenges presented by current highly biomimetic robotic arms. Concurrently, it offers an opportunity to corroborate findings from classical anatomical studies. The skeletal model adopted for this robotic arm utilizes commercially available 3D scanning files, and the biological soft tissues are substituted with suitable engineering materials.

\subsection{The design of the soft tissues}
\subsubsection{\textbf{Glenoid labarum}}

\begin{figure*}[htb]
\centerline{\includegraphics[width=1\textwidth]{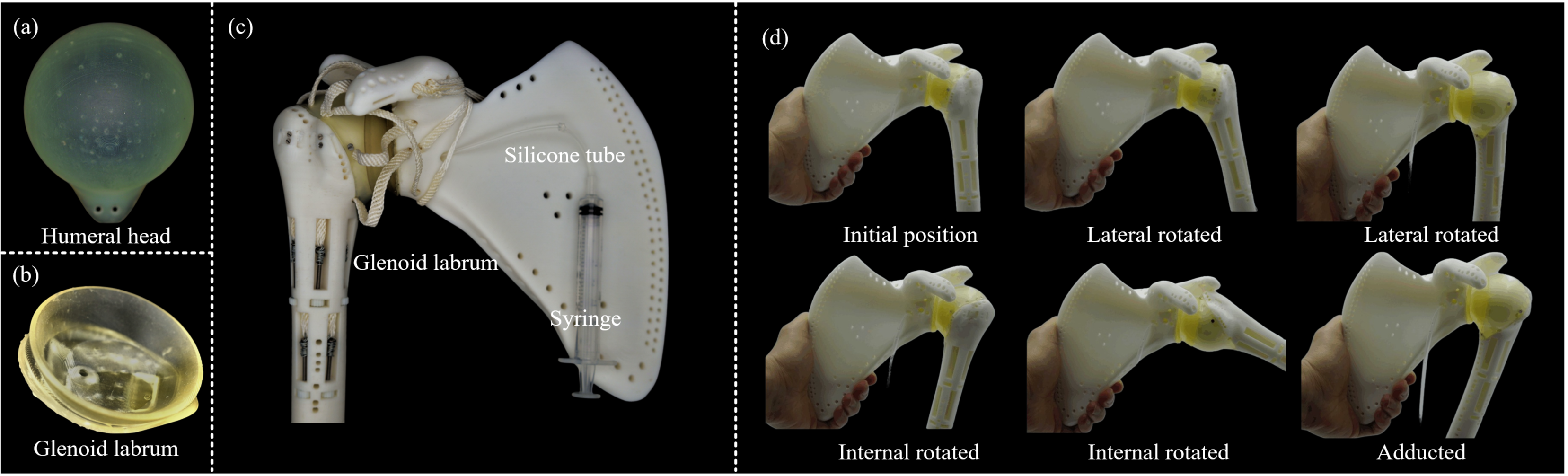}}
\caption{(a) Humeral head cartilage; (b) Glenoid labrum; (c) Negative intra-articular pressure formed by the humeral head and glenoid labrum; (d) The glenohumeral joint with lubricating oil during flexion/extension, rotation, adduction/abduction movements.}
\label{fig5.9}
\end{figure*}

Researchers have discovered that the glenoid labrum can act as an anti-shear bumper, and this characteristic has been replicated in the design of the glenohumeral joint. As illustrated in Fig.\ref{fig5.9}(c), the humeral head is coated with a 1.5mm thick cartilage (Fig.\ref{fig5.9}(a)), which is printed with Formlabs durable resin {(Elongation at break: 55\%, Ultimate Tensile strength: 28 MPa, Tensile modulus: 1 GPa)} using Formlabs Form 3 printer in precision mode (0.025mm/layer). This high precision enables a smooth and polished finish. The glenoid labrum (Fig.\ref{fig5.9}(b)) is printed using Formlabs elastic resin, which has similar properties to silicone and allows for airtight adhesion to the humeral head. The glenoid labrum has a vent hole that can be connected to a syringe through a silicone tube. By applying lubricating oil to the humeral head and attaching the glenoid labrum, the air inside the labrum is extracted using the syringe, creating a negative pressure between the humeral head and the labrum. This negative pressure allows the humeral head to be securely attached to the labrum, providing stability to the glenohumeral joint. 
{Upon evaluation, it was found that the lubricating oil effectively mitigated excessive damping in the humerus during flexion/extension, rotation, and adduction/abduction movements, ensuring smooth operation when equipped with artificial muscles}, as shown in Fig.\ref{fig5.9}(d). The negative pressure generated between the humeral head and the labarum provided stable adhesion and was capable of withstanding considerable tension (more than 50N).

\subsubsection{\textbf{Ligaments}}

However, it was found experimentally that the flexible material of the glenoid labrum tends to deform slightly after prolonged joint motion, leading to a reduction in air tightness. Therefore, a pre-tensioned ligament system was developed to provide stable joint fixation and avoid the need for constant manual lubrication while maintaining the same performance. The function of the negative pressure chamber is thus eliminated and supplanted by pre-tensioned ligaments.

\begin{figure*}[htb]
\centerline{\includegraphics[width=1\textwidth]{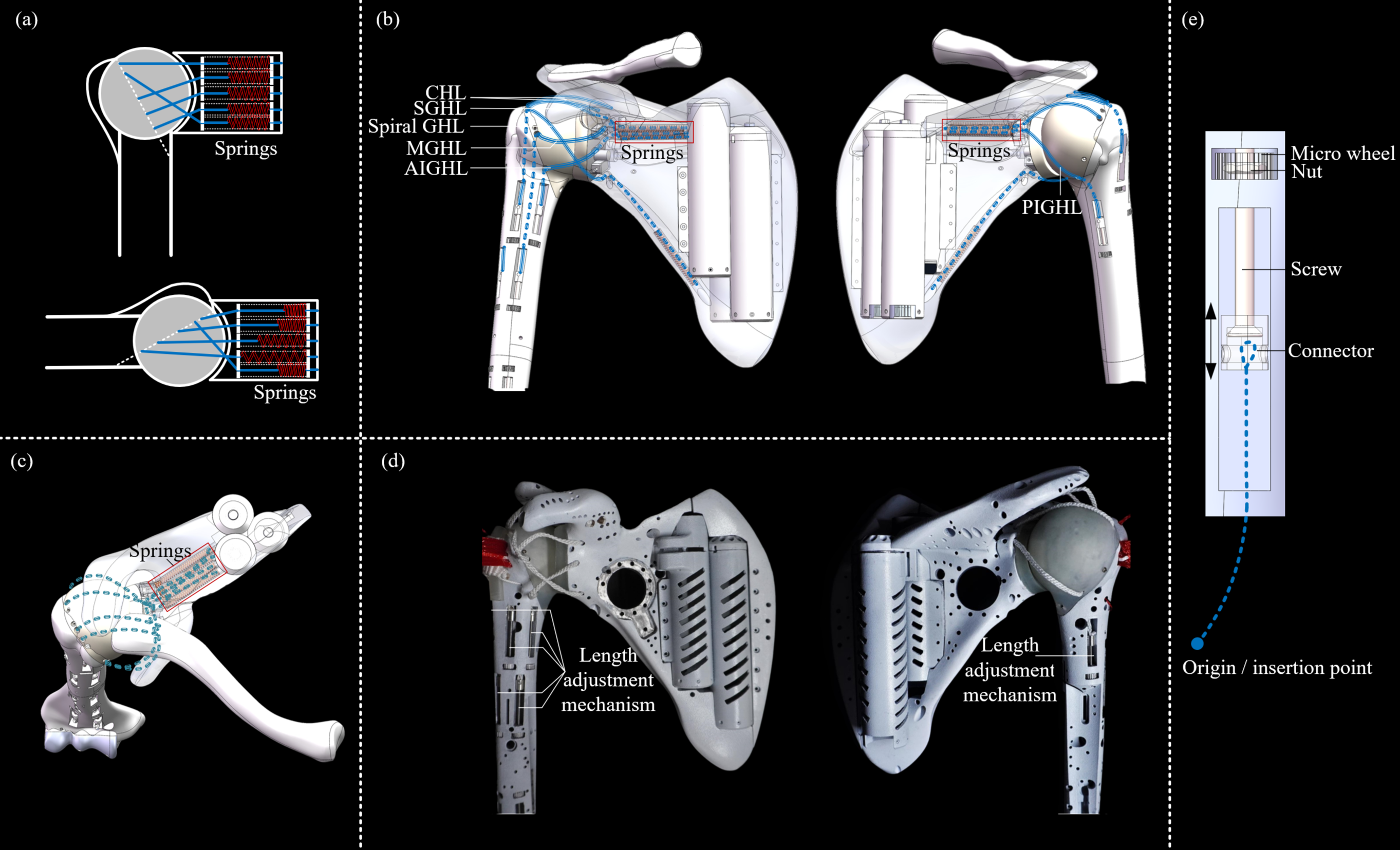}}
\caption{(a) Simplified diagram of the pretensioned ligament system; (b) Arrangement of ligaments in the glenohumeral joint. (c) Top view of six ligaments with the pre-tension mechanism. (d) The arrangement of the pre-tensioned ligament system and length adjustment mechanisms in the prototype. (e) Detail of the length adjustment mechanism. }
\label{fig5.10}
\end{figure*}

According to anatomy, the ligaments in the glenohumeral joint can be simplified into seven major portions to limit the joint's position. These seven ligaments were replicated on the prototype based on the approximate location of their origin and insertion points, as indicated in {blue} in Fig.\ref{fig5.10}(b)(c). The ligament, {composed of six polyethene fishing wires (each with a diameter of 0.55 mm and breaking strength of 45.36 kg), is pre-stretched with 200 N force prior to installation to approximate a non-extensible length.} Due to the elimination of the glenoid labrum with negative pressure and the overlong length of the ligaments required in the joint's initial position (too short would reduce the range of motion of the joint), the ligaments are in a state of extreme laxity and unable to stabilize the joint. To provide the joint with basic stability similar to negative pressure, i.e., stability even without the contribution of tendons, built-in compression spring systems are used. This design applies a {10 N preload to each ligament}, allowing them to be tensioned at the joint's initial position, as shown in the diagram in Fig.\ref{fig5.10}(a). During the dynamic process of joint articulation, the alterations in ligament length consequently engender more significant deformation in the associated spring mechanism. As shown in Fig.\ref{fig5.10}(b) and (c), except for CHL, one end of the ligaments is fixed into the scapula through compression springs (inside the scapula, {represented and boxed in red}). When the ligaments are tensioned, the spring is compressed, and the exposed ligaments will be extended. The alternate ends of the ligament bundles insert into the humeral head, traverse the humerus internally, and anchor to the length adjustment mechanisms (Fig.\ref{fig5.10}(d)). The operative principle of these mechanisms is illustrated in Fig.\ref{fig5.10}(e). The ligament bundles connect to the mechanism's connectors via the humerus's internal conduit. Rotating the micro-wheel induces axial movement of adjustment screws within the slots, modulating the length of the ligaments. This mechanism permits a ligament length adjustment up to 20mm.

The length of the ligaments is determined through testing during the development process. The initial length of the ligaments is estimated and then adjusted using the above-mentioned mechanism. The length of the ligaments is adjusted until the springs are compressed to their solid length when the joint moves to the limited position of each degree of freedom and the exposed ligaments cannot be extended any further. This helps to limit the position of the joint. Due to the limited elastic travel of the spring, once the ligaments have been adjusted to the appropriate length, the exposed length of the ligaments will resume to the minimum when the joint returns to its initial position, which may be either tensioned or relaxed. During the movement of the joint, the ligaments will be strengthened and springs will be compressed, which provides resistance to the joint dislocation and thus provide basic stability to the joint. Due to the low coefficient of elasticity of the compression springs used, this mechanism does not cause excessive resistance to joint movement. Compared to the design without the spring, in this design, the length of the ligament is directly set to the desired maximum length. However, when the joint is in its initial position, the ligaments are not tensioned and do not provide any stability to the joint. This may result in easy dislocation of the joint in the absence of any tendon contribution.


\subsubsection{\textbf{Tendon and muscles}}
\begin{figure*}[htb]
\centerline{\includegraphics[width=1\textwidth]{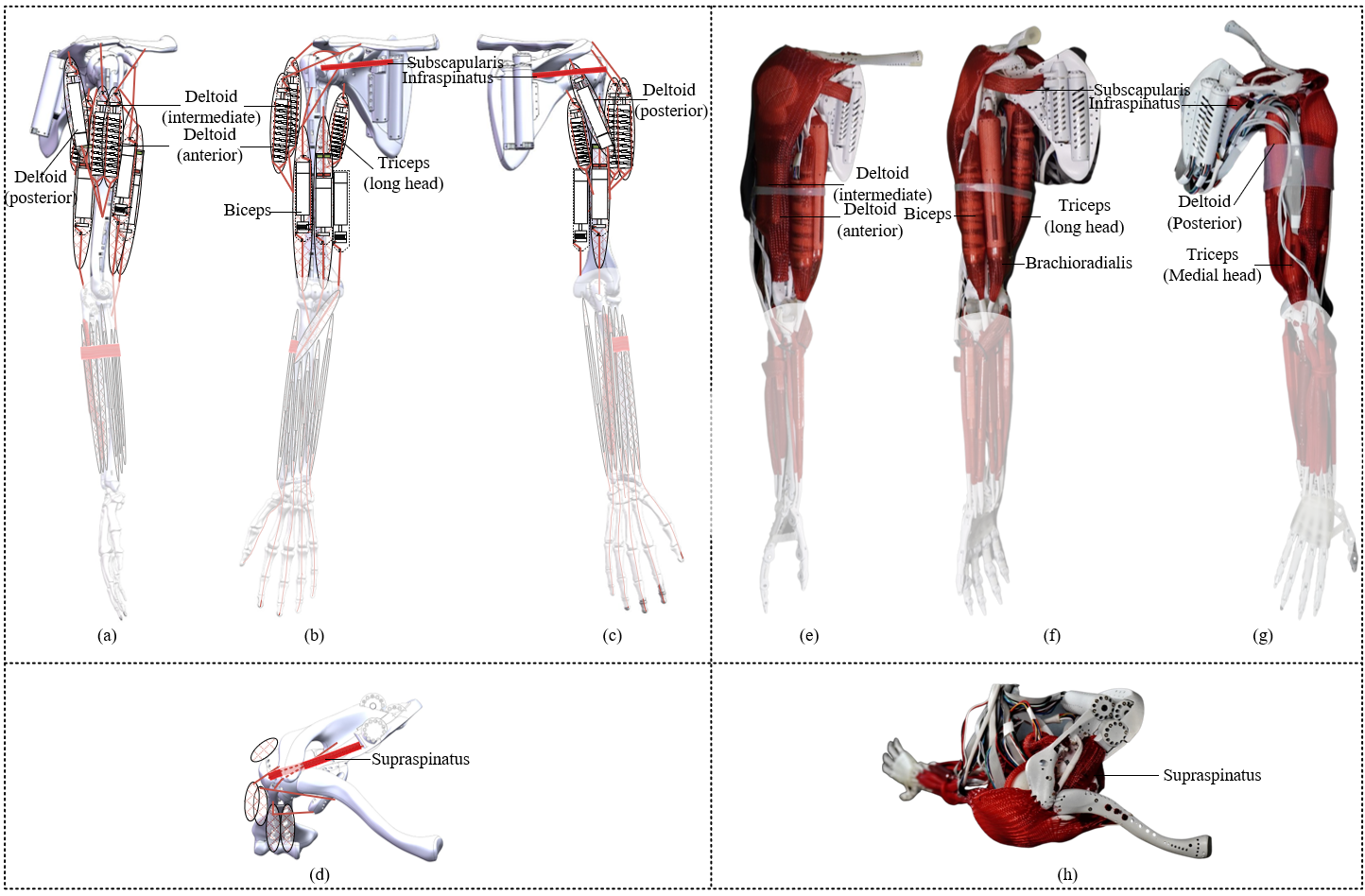}}
\caption{Tendon and muscle arrangement of the proposed robotic shoulder: (a) Right view; (b) Front view; (c) Rear view; (d) Top view. Tendon and muscle arrangement of the proposed robotic shoulder prototype. (a) Right view; (b) Front view; (c) Rear view; (d) Top view}
\label{fig5.13}
\end{figure*}
To address the size and weight limitations, 7 major muscles are reproduced on the robotic shoulder, as listed in Table \ref{tab5.2}. The muscles are powered by 4 soft actuators, comprising four ECAs (External Spring Compliant Actuators). The other muscles are actuated using non-compliant actuators. Fig. \ref{fig5.13}(a)-(d) illustrates the distribution of the muscles, including the motors and tendons. The motors and driving pulleys for the subscapularis, infraspinatus, and supraspinatus muscles are embedded within the scapula due to space constraints. In the proposed prototype, the tendon is covered by a red braided sleeve, as shown in Fig.\ref{fig5.13}(e) to (h). 

\begin{table}[htb]
\caption{Artificial muscles applied in the proposed robotic arm and hand.}
\footnotesize
\begin{center}
\begin{tabular}{ l c}
\toprule
 Muscle & Actuator Type\\
\midrule
Deltoid (anterior)     & ECA   \\
Deltoid (intermediate) & ECA  \\
Deltoid (posterior)    & ECA    \\
Subscapularis          & Without spring       \\
Infraspinatus          & Without spring        \\
Supraspinatus          & Without spring        \\
Triceps (Long head)    & ECA     \\
\bottomrule
\end{tabular}
\label{tab5.2}
\end{center}
\end{table}

\section{Characteristics of the Glenohumeral Joint}
\subsection{Incomplete ball and socket structure in the glenohumeral joint}

To enable a wide range of motion, the glenohumeral joint's ligaments only tense as the joint near its extreme positions. Joint stability primarily derives from the cross-joint tendons, such as the deltoid, biceps, and rotator cuff. This mechanism efficiently mitigates the stability-mobility trade-off. As the humeral angle alters and force direction coincides with a tendon, the tendon can resist the tension directly, thus fortifying the glenohumeral joint. The simplicity of this mechanism obviates the need for mathematical analysis. Discussion is limited to situations of natural shoulder adduction under axial force.

\begin{figure}[htb]
\centerline{\includegraphics[width=0.65\textwidth]{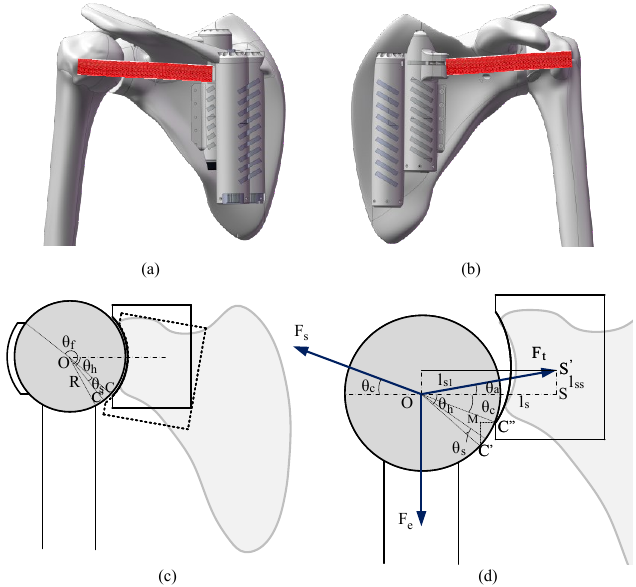}}
\caption{Rotator cuff in the glenohumeral joint: (a) rear view (b) front view. (c) The simplified diagram of the glenohumeral joint (humeral head and glenoid). (d) The force analysis diagram of the ball and socket structure when the glenohumeral joint dislocated.}
\label{fig6.18}
\end{figure}

Fig. \ref{fig6.18}(c) illustrates a simplified diagram of the glenohumeral joint. When the humerus is naturally adducted, the humeral head and glenoid of the scapula form an incomplete ball and socket joint. The edge of the joint contact surface is denoted as $C$ when the angle between the scapula and the horizontal is 0 (shown by the solid line). However, when the angle between the scapula and the horizontal is $\theta_s$, as shown by the dashed line, the contact edge point changes to $C'$. The angle between $C'O$ and the horizontal is $\theta_h$. As $\theta_s$ increases, $\theta_h$ also increases, resulting in increased joint stability when a force is applied in the vertical direction.

Assuming an axial force $F_e$ is applied to the humerus, as shown in Fig. \ref{fig6.18}(d), the glenohumeral joint is subjected to a tendency to dislocate. At this point, the situation can be simplified by considering three forces acting on the humeral head. The first is the applied force $F_e$ itself, while the other two are the support force $F_s$ from the joint contact point, and the tendon force $F_t$ from the infraspinatus and supraspinatus. However, the frictional force that further prevents joint dislocation is not considered in this simplified model.

When the joint is dislocated, the joint contact point slides from $C'$ to $C''$. The length of the tendon $OS$ (to represent the tendons of infraspinatus and supraspinatus) is stretched from $l_{s}$ to $l_{s1}$. The position of tendons changes from $OS$ to $OS’$, and the angle with the horizontal line increases to $\theta_a$. $\theta_h$ decreased to $\theta_c$.

According to the force balance, there are:
\begin{equation}
\begin{cases}
F_tsin\theta_a+F_ssin\theta_c=F_e\\
F_tcos\theta_a=F_scos\theta_c
\end{cases}
\label{eq6.33}
\end{equation}

In $\Delta OSS’$, it has:
\begin{equation}
\begin{cases}
l_{ss}=R(sin\theta_h-sin\theta_c) \\ 
l_s=l_{t0}+R(cos\theta_c-cos\theta_h) \\
l_{s1}=\sqrt{l_s^2+l_{ss}^2\ }\\
cos\theta_a=\frac{l_s}{l_{s1}} \\
sin\theta_a=\frac{l_{ss}}{l_{s1}}
\end{cases}
\label{eq6.35}
\end{equation}

Where, $l_{ss}$ is $SS'$, and it is equal to $C'M$. $R$ is the radius of the humeral head. $l_{s}$ is $OS$, $l_s=l_{t0}+MC''$, $l_{t0}$ is the initial length of $OS$, which is known. $l_{s1}$ is $OS'$. 

When the motor keeps still, $F_t$ can be calculated:
\begin{equation}
F_t=k_{t}(l_{s1}-l_{t0})
\label{eq6.36}
\end{equation}

Where $k_t$ is the stiffness of the tendon.

Combine equations \ref{eq6.33} to \ref{eq6.36}, the relation between $F_e$ and $\theta_c$ in different $\theta_h$, $F_e=f_{ch}(\theta_c,\theta_h)$ can be obtained.

The angle $\theta_h$ can be adjusted by modifying the joint contact surface on the scapula, such as changing the size of the socket in the ball and socket structure of the glenohumeral joint, or by rotating the scapula and altering $\theta_s$ to adjust the joint contact surface.

\begin{figure}[htb]
\centerline{\includegraphics[width=0.5\textwidth]{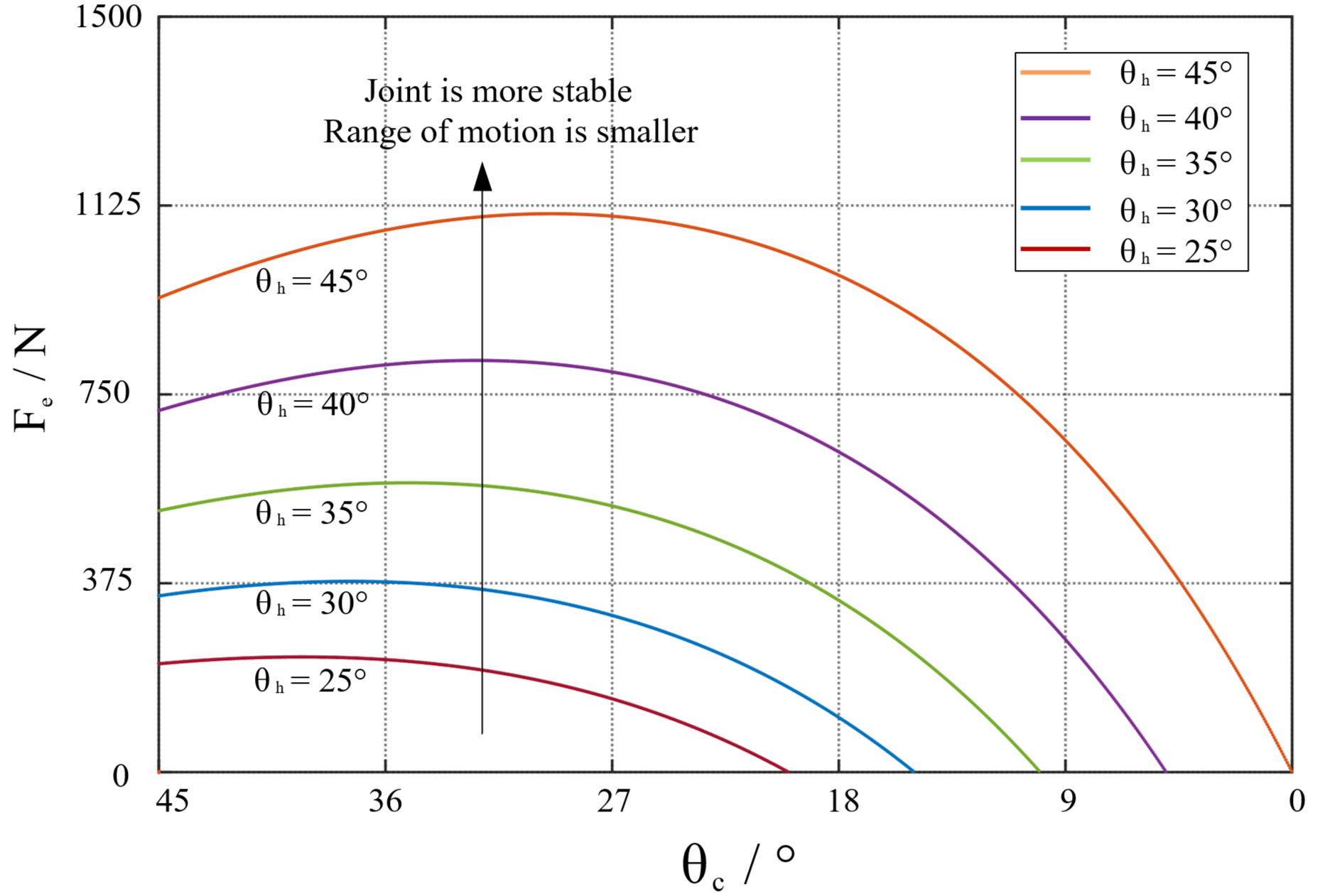}}
\caption{The simulation result of $F_e=f(\theta_c,\theta_h)$ in different $\theta_h$.}
\label{fig6.19}
\end{figure}

Fig. \ref{fig6.19} presents $F_e=f(\theta_c)$ in different $\theta_h$. The blue curve presents the situation when $\theta_h=30\degree$. In this setting, if the external force $F_e$ reaches $400N$ and does not withdraw, the joint will be completely dislocated until $\theta_c$ decreases to 0. The maximum axial external force the joint can withstand is $400N$ when $\theta_h=30\degree$. It is observable that with an increase in $\theta_h$, the joint's resistance to the maximum value of $F_e$ correspondingly elevates, thereby enhancing the joint stability.

\subsection{Coupling stability of humeroradial and glenohumeral joints}

The long head of the biceps muscle crosses both the humeroradial and glenohumeral joints, potentially increasing stability in the glenohumeral joint when the biceps actuate elbow flexion. This coupling of joint stability allows the two-joint system to avoid significant shortcomings due to reduced load capacity in one joint.

\begin{figure}[htb]
\centerline{\includegraphics[width=0.65\textwidth]{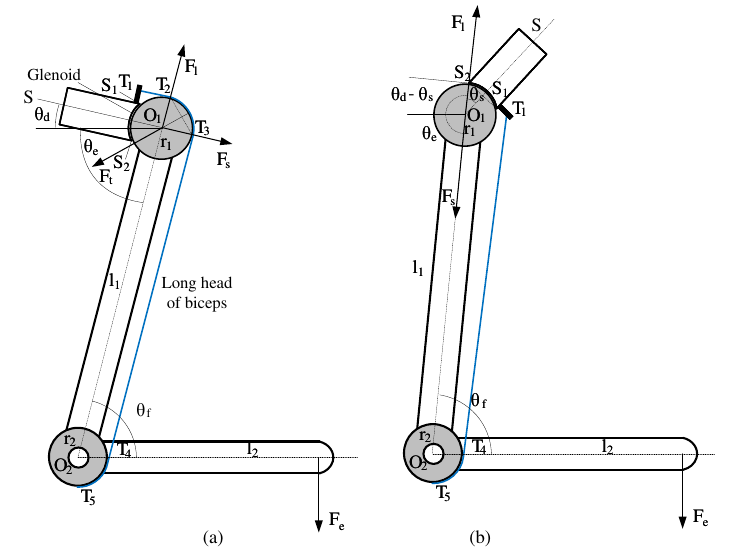}}
\caption{(a) Simplified diagram of the glenohumeral joint and elbow when the long head of the biceps is retained. (b) A condition that stability coupling failure.}
\label{fig6.21}
\end{figure}

Fig. \ref{fig6.21} depicts the simplified diagram of the glenohumeral joint, illustrating the isolated long head of the biceps in the absence of other soft tissues such as ligaments and tendons. When an external force $F_e$ is applied to the distal end of the forearm, the tension in the biceps increases. As the long head of the biceps also crosses the glenohumeral joint, it can press the humeral head into the glenoid, resulting in a more stable ball and socket joint.

Assuming that the length of the long head of the biceps tendon (shown in blue in Fig. \ref{fig6.21}(a)) remains constant, the angle between the two joints, $\theta_e$ and $\theta_f$, are coupled. The relationship between them is determined by the radius of the humeral head, $r_1$, and the moment arm of the elbow, $r_2$, and can be expressed as:
\begin{equation}
\left(\theta_d+\theta_e\right)r_1=\theta_f r_2
\label{eq6.37}
\end{equation}

Where $\theta_d$ is the angle between the perpendicular bisector of the line $S_1S_2$ ($S_1$ and $S_2$ are the edge points of glenohumeral joint contact surface), denoted as $O_1 S$, and the horizontal line. $\theta_d$ is known. $\theta_e$ is the angle between the axial line of the humerus and the horizontal line. $\theta_f$ is the angle between the axial line of the forearm and the humerus.

When a vertical downward external force $F_e$ is applied to the distal end of the forearm, the tension in the tendon will increase. The whole system will stabilise at a position. It should be the position that has the lowest potential energy, where $\theta_e$ and $\theta_f$ lead $H$ at its minimum value:
\begin{equation}
H=l_{1} sin\theta_{e}+l_{2} \sin(\theta_{f}-\theta_{e})
\label{eq6.38}
\end{equation}

Where, $l_1$ is the distance between $O_1$ and $O_2$. $l_2$ is the distance between $O_2$ and the force acting point. Combine equation \ref{eq6.37}, $\theta_e$ can be calculated.

The force analysis of the humeral head is conducted in the equilibrium position as shown in Fig. \ref{fig6.21}(a). Without considering friction and gravity, it can be simplified to three forces acting on the humeral head. The first force is the support force $F_l$ along the axial line of the humerus. The second force is the combined force $F_t$ due to the tendon warping the humeral head. The direction of $F_t$ is the perpendicular bisector of $T_2T_3$. $T_2$ and $T_3$ are contact points of tendon $T_1T_2$ and $T_3T_4$ with the humeral head. $F_t$ across the centre of the humeral head $O_1$. The third force is the support force $F_s$ to the humeral head (ball) from the glenoid of the scapula (socket). The direction of $F_s$ across $O_1$, and should be balanced with the other two forces. In the position shown in Fig. \ref{fig6.21}(a), the direction of $F_s$ is at the perpendicular bisector of $S_1S_2$ ($S_1$ and $S_2$ are the contact edge points between glenoid of scapula and humeral head).

It is assumed that the tendon is unstretchable and will not break and the strength of the skeletal material is infinite. When $\theta_d-\theta_s+\theta_e<180\degree$ ($\theta_s$ is the angle of $\angle S_2 O_1 S$), even if $F_e$ increases infinitely, the three forces above remain balanced and the system remains stable. If the tendon can be stretched but will not break, the stretched amount may only cause $\theta_e$ and $\theta_f$ to be balanced at the new position and the system may remain stable. These are the conditions to realize stability coupling in the elbow and shoulder due to the biceps crossing both joints.

$\theta_d+\theta_e-\theta_s=180\degree$ is the condition for stability coupling failure, where $O_2$, $O_1$, $S_2$ are in the same line. The joint contact surface between the scapula and the humeral head $S_1S_2$ can not provide the splitting force that is perpendicular to $O_1O_2$ to the right in order to prevent joint dislocating. The humeral head will tend to dislocate to the left as shown in Fig. \ref{fig6.21}(b). Thus, $\theta_d-\theta_s+\theta_e<180\degree$ is the condition to maintain glenohumeral joint stability if only the long head of the biceps is retained. When $\theta_d+\theta_e$ increases, the stability of the glenohumeral joint will decrease.

\subsection{The self-locking mechanism}

\begin{figure}[htb]
\centerline{\includegraphics[width=0.65\textwidth]{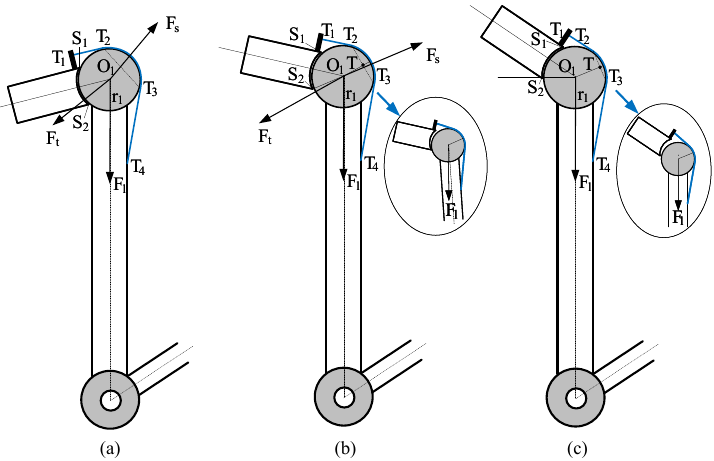}}
\caption{(a) Self-locking mechanism is in active in the glenohumeral joint; (b) Self-locking mechanism fails (or reduces joint stability) in the glenohumeral joint. (c) The threshold for the self-locking mechanism fails.}
\label{fig6.22}
\end{figure}

When the arm is naturally adducted, the tendons cross the glenohumeral joint (deltoid, subscapularis, biceps) and incorporate with the ball and socket structure in the glenohumeral joint forming a self-lock mechanism. The subscapularis functions in a similar way to the biceps, which wraps around the humeral head. As shown in Fig. \ref{fig6.22}(a), $T_1T_2T_3T_4$ represents the tendon. When an axial load $F_1$ is applied to the humerus, the combined force from the tendon wrapping the humeral head is $F_t$. There will be a support force $F_s$ from the glenoid, its direction may be across the joint contact edge point $S_2$. When the joint is in the position shown in  Fig. \ref{fig6.22}(a), these three forces should be balanced and the self-lock mechanism is activated. As demonstrated in Fig. \ref{fig6.22}(b), the system stability diminishes during scapula rotation, leading to potential joint dislocation as a consequence of tendon stretching. Without this tendon stretching, the humeral head may slip from the glenoid under minor humerus rotation, as illustrated in the enclosed diagram. The threshold for self-locking mechanism failure may reached when point $S_2$ is higher than $O_1$, as shown in Fig. \ref{fig6.22}(c), and the external force will stretch the tendon directly and dislocate the joint slightly, as shown in the enclosed diagram.

\section{Static Analysis and Simulation}

In this section, the relation between the glenohumeral joint's output torque for each motion and its positional parameters will be established and simulated. This analysis is necessary because, in a high-fidelity design, joint movements often induce alterations in tendon force directions.

\subsection{Glenohumeral joint flexion/extension}

\begin{figure*}[htb]
\centerline{\includegraphics[width=1\textwidth]{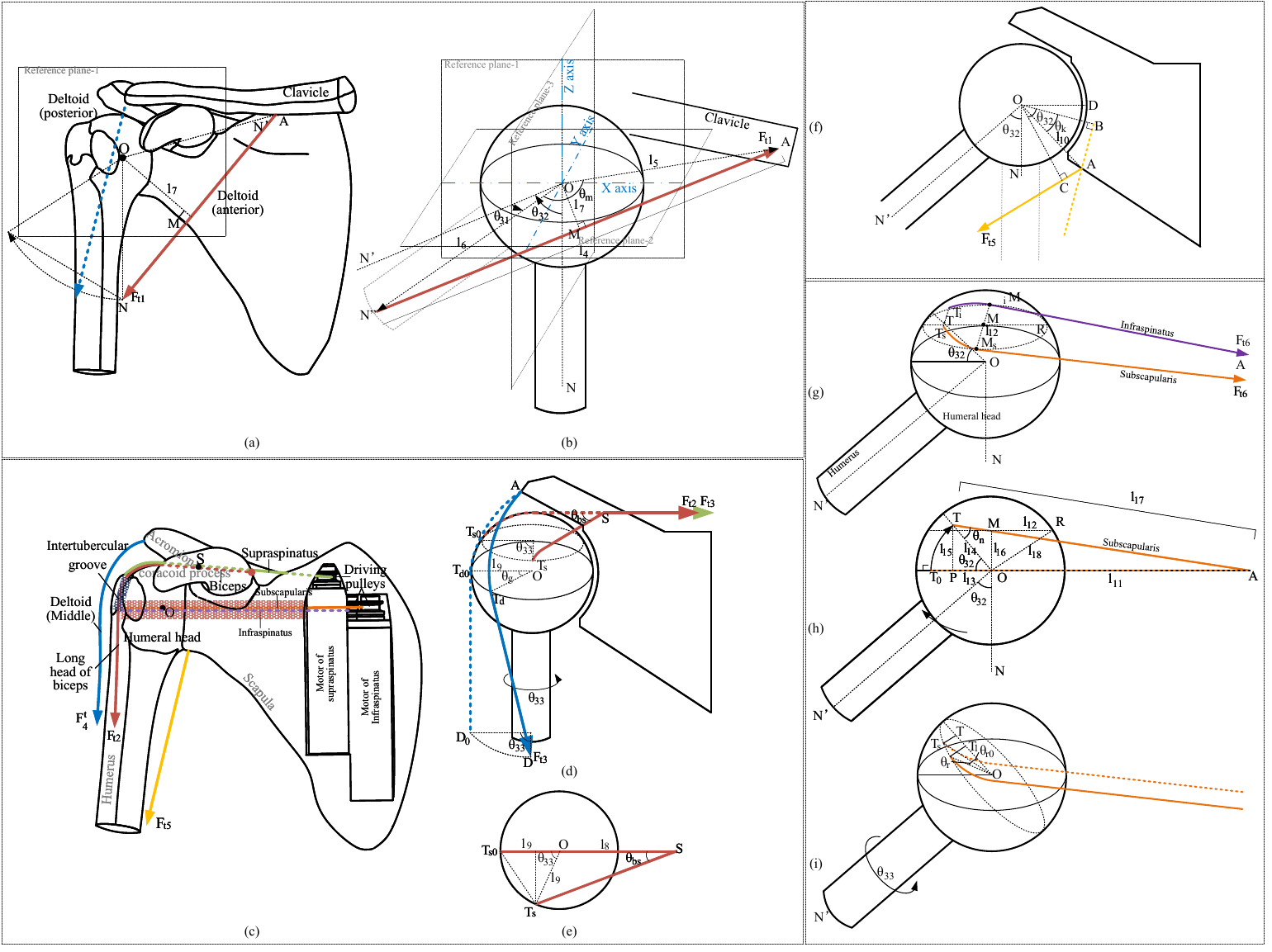}}
\caption{(a) Actuation system for glenohumeral joint flexion/extension; (b) Simplified schematic of flexion/extension actuation; (c) Actuation system for glenohumeral joint abduction/adduction; (d) Simplified schematic of abduction actuation; (e) Top view of abduction actuation system; (f) Simplified diagram of adduction actuation system for glenohumeral joint; (g) Simplified structure for glenohumeral joint rotation actuation; (h) Cross-sectional view of joint rotation actuation system; (i) Status of actuation system during humeral rotation.}
\label{fig7.10}
\end{figure*}

In the robotic arm prototype, the glenohumeral joint flexion is achieved by the deltoid (anterior) muscle, since the torso is absent. The tendon of the deltoid (anterior) originates from the lateral surface of the humerus and inserts into the middle section of the clavicle, as shown by the red line in Fig. \ref{fig7.10}(a). As the humerus rotates, the movement of the origin point of the deltoid is minor and has little effect on the joint output torque. Therefore, only the relationship between the joint output torque $\tau_{31f}$ (glenohumeral joint flexion), the joint abduction angle $\theta_{32}$, and the joint flexion angle $\theta_{31}$ will be analyzed in this case.

This structure can be simplified as shown in Fig. \ref{fig7.10}(b). To facilitate the calculation, a coordinate system is created as shown in the figure. As the humerus moves from the initial position $ON$, it abducts $\theta_{32}$ around the y-axis to position $ON'$, and then flexes $\theta_{31}$ around the z-axis to $ON''$. 

$\overrightarrow{ON}$ after rotation to $\overrightarrow{ON''}$ is:
\begin{equation}
\overrightarrow{ON''}=(-l_6\cos(\frac{\pi}{2}-\theta_{32})cos\theta_{31},l_6sin\theta_{31},-l_6\sin(\frac{\pi}{2}-\theta_{32}))
\label{eq7.22}
\end{equation}

Where, $l_6$ is the length of $ON''$, and it is constant. 

$\cos{\theta_m}$ can be calculated as:
\begin{equation}
\cos{\theta_m}=\frac{\overrightarrow{OA}\cdot\overrightarrow{ON''}}{l_5l_6}
\label{eq7.21}
\end{equation}

Where $\theta_{m}$ is the angle of $\angle AON''$, $l_5$ is the length of $OA$, and it is constant.

According to the cosine theorem, $l_4$ is obtained:
\begin{equation}
l_4=\sqrt{l_5^2+l_6^2-2l_5l_6cos\theta_m}
\label{eq7.20}
\end{equation}

The length of $OM$ (perpendicular to $AN''$), $l_7$ can be calculated as:
\begin{equation}
l_7=\frac{l_5l_6\sin\theta_m}{2l_4}
\label{eq7.19}
\end{equation}

The joint torque can be calculated as:
\begin{equation}
\tau_{31f}=F_{t1} l_7
\label{eq7.18}
\end{equation}

\begin{figure*}[htb]
\centerline{\includegraphics[width=1\textwidth]{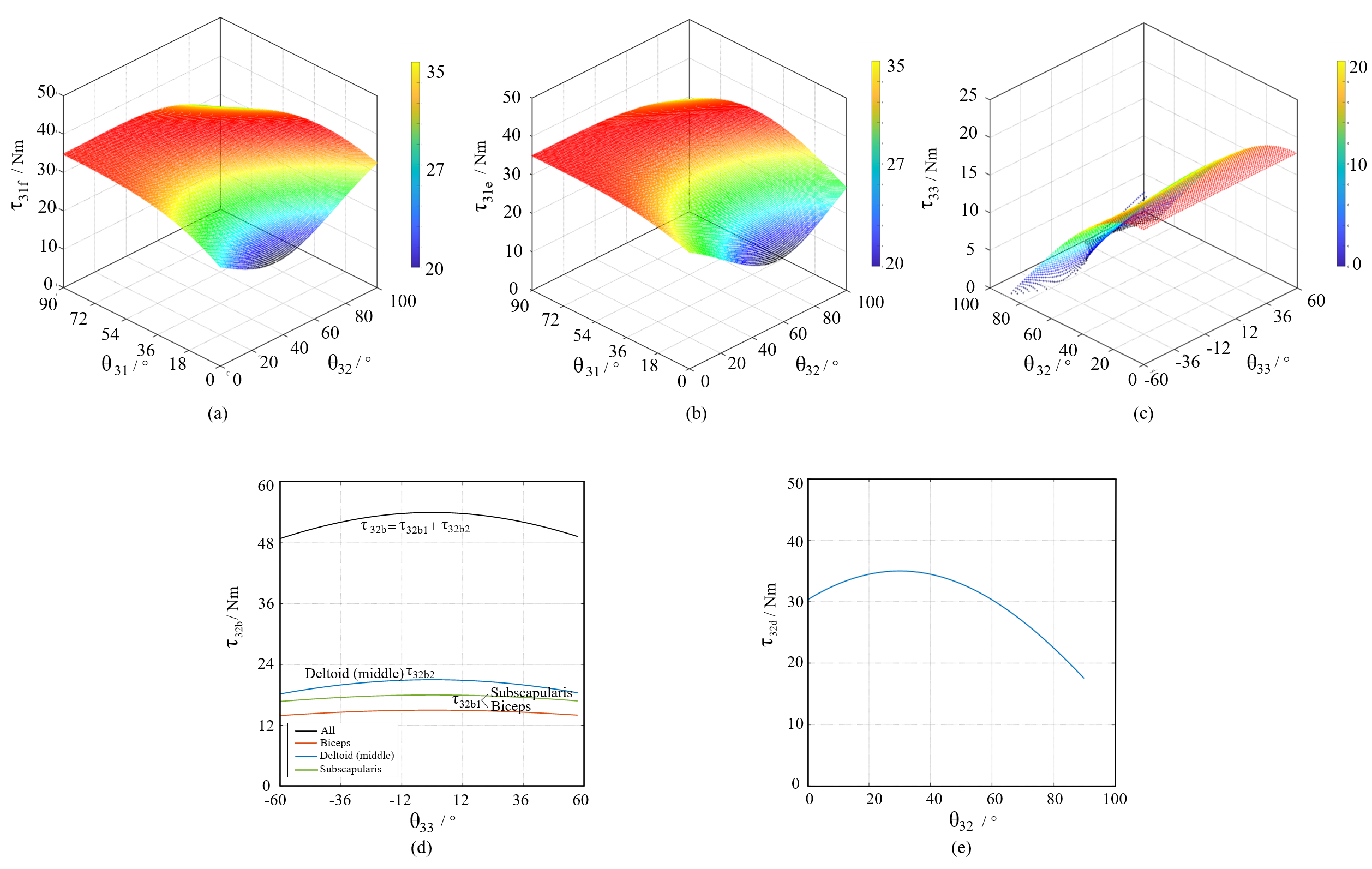}}
\caption{The simulation result of the relation between (a) $\tau_{31f}$, $\theta_{32}$ and $\theta_{31}$, (b) $\tau_{31e}$, $\theta_{32}$ and $\theta_{31}$, (c) $\tau_{33}$ and $\theta_{32}$, $\theta_{33}$, (d) $\tau_{32b}$ and $\theta_{33}$ for the three muscles, (e) $\tau_{32d}$ and $\theta_{32}$.}
\label{fig7.11}
\end{figure*}

Leveraging a Maxon EC 4-pole 120W motor with a 1:128 gearbox enables the establishment of the relationship between $\tau_{31f}$ and $\theta_{32}$, $\theta_{31}$ when $F_{t1}$ is equal to 700N. This relationship is depicted in Fig. \ref{fig7.11}(a).

In the proposed robotic arm, the deltoid (posterior) is responsible for driving the glenohumeral joint extension. The blue line in Fig. \ref{fig7.10} shows the origin of the deltoid (posterior) from the lateral surface of the humerus and its insertion point on the spine process of the scapula. By applying the same calculation approach, the relationship between the glenohumeral joint extension output torque $\tau_{31e}$ and joint angles $\theta_{32}$ and $\theta_{31}$ can be determined, as shown in Fig. \ref{fig7.11}(b).

\subsection{Glenohumeral joint abduction}

In the robotic arm prototype, the glenohumeral joint abduction can be driven by three muscles, namely the deltoid (middle), the supraspinatus, and the long head of the biceps. As illustrated in Fig. \ref{fig7.10}(c), the motor of the biceps is positioned on the lateral aspect of the middle part of the humerus, the red-coloured tendon passes through the intertubercular groove and crosses the notch between the acromion and coracoid process on the scapula before inserting into the scapula. The motor of the supraspinatus is located inside the scapula, the green-coloured tendon crosses the groove between the acromion and coracoid process and connects to the protrusion on the side of the humeral head. 

The paths of the supraspinatus and the long head of the biceps are in close proximity and can be discussed together. The long head of the biceps crosses the 'tube' formed by the intertubercular groove (Fig. \ref{fig7.10}(c)). As the humerus rotates, the tendon insertion point $T_{s0}$ of both muscles will also rotate to $T_s$, as shown by the red line in Fig. \ref{fig7.10}(d). The red tendon will move from its dashed line position to the solid line position by an angle $\theta_{33}$. As the tendon crosses the notch between the acromion and coracoid process, it bends at an angle $\theta_{bs}$ at point $S$. According to the top view in Fig. \ref{fig7.10}(e), $\theta_{bs}$ can be calculated as:
\begin{equation}
cos\theta_{bs}=\frac{l_8+l_9cos\theta_{33}}{\sqrt{\left(l_8+l_9cos\theta_{33}\right)^2+{(l_9 sin\theta_{33})}^2\ }}
\label{eq7.14}
\end{equation}

Where $l_8$ is the length of $OS$, $L_9$ is the length of $OT_{s}$, as shown in Fig. \ref{fig7.10}(e).

The output torque when the supraspinatus and the long head of the biceps drive the glenohumeral joint abduction can be calculated approximately as:
\begin{equation}
\tau_{32b1}=\left(F_{t2}+F_{t3}\right)l_9cos\theta_{bs}
\label{eq7.15}
\end{equation}

Where $F_{t2}$ and $F_{t3}$ represent the tension force generated by the supraspinatus and biceps, respectively.

The motor of the deltoid (middle) is located on the lateral side of the humerus, with the tendon (blue) attached to the acromion of the scapula. As the humerus rotates, the origin point of the tendon on the motor, $D_0$, will move from the dashed to the solid position to $D$, as shown in Fig. \ref{fig7.10}(d). The contact point between the tendon and the humeral head moves from $T_{d0}$ to $T_{d}$. Due to the complexity of the structure, $\theta_g$ (the angle of $\angle$$T_{d0} O T_{d}$) will vary as the joint abducts, approximated as $\theta_g=0.5\theta_{33}$. The output torque when the deltoid (middle) drives the joint to abduct, $\tau_{32b2}$ can be approximately calculated as:
\begin{equation}
\tau_{32b2}=F_{t4}l_9cos\theta_g
\label{eq7.16}
\end{equation}

Where $F_{t4}$ represent the tension force generated by the deltoid (middle).

Given that the supraspinatus employs a Maxon ECX TORQUE 22mm motor with a 1:62 gearbox, the biceps uses a Maxon EC 4-pole 90W motor with a 1:128 gearbox, and the deltoid (middle) incorporates a Maxon EC 4-pole 120W motor with a 1:128 gearbox, the relationship between the output torque, $\tau_{32b}$ ($\tau_{32b}=\tau_{32b1}+\tau_{32b2}$), and $\theta_{33}$ can be established when three tendons, with forces $F_{t2}$ = 600 N, $F_{t3}$ = 500 N, and $F_{t4}$ = 700 N respectively, drive the joint abductions. These relationships are illustrated in Fig. \ref{fig7.11}(d). The deltoid (middle) contributes the most to $\tau_{32b}$. The glenohumeral joint rotation angle does not exceed $\pm60\degree$ and has a minor effect on the output torque $\tau_{32b}$. If all three tendons act simultaneously, the joint torque can reach a maximum of 54 Nm.

\subsection{Glenohumeral joint adduction}

In the robotic arm prototype, the glenohumeral joint adduction is actuated by the long head of the triceps, as shown by the yellow line in Fig. \ref{fig7.10}(c). The motor of the long head of the triceps is not rigidly fixed to the humerus, and therefore, the position of the tendon is less displaced when the humerus rotates, remaining approximately in its original position (i.e., the position marked in yellow in Fig. \ref{fig7.10}(c)). Thus, only the relationship between the output torque $\tau_{32d}$ and the adducted angle $\theta_{32}$ is considered. The structure can be simplified as shown in Fig. \ref{fig7.10}(f). The output torque $\tau_{32d}$ for glenohumeral joint adduction can be calculated as:
\begin{equation}
\tau_{32d}=F_{t5}l_{10}\cos{(\theta_k-\theta_{32})}
\label{eq7.17}
\end{equation}

Where, $l_{10}$ is the distance from the insertion point $A$ of the tendon in the scapula to the centre $O$ of the humeral head. $F_{t5}$ is the tendon force output by the long head of the triceps. $\theta_k$ is the angle between the moment arm $OB$ and $OA$ when the joint is in its initial position. Taking in the position parameters $l_{10}$ and $\theta_k$ of the prototype, given $F_{t5}$ = 700 N (Maxon EC 4-pole 22mm 120W motor with 1:128 gearbox is used to drive the long head of triceps), the relationship between $\tau_{32d}$ and $\theta_{32}$ can be obtained as shown in Fig. \ref{fig7.11}(e).

\subsection{Glenohumeral joint rotation}

In the robotic arm prototype, the glenohumeral joint rotation is driven by the subscapularis and infraspinatus, as shown in Fig. \ref{fig7.10}(c) (subscapularis is represented by the orange line, infraspinatus is represented by the purple line). The motors of these two muscles are located inside the scapula, and their tendons are connected to the humeral head. The subscapularis assists in internal rotation, and the infraspinatus drives external rotation. The two muscles are arranged symmetrically. When the glenohumeral joint is in its initial position, as shown in Fig. \ref{fig7.10}(c), the tendon passes through the centre of the circle of the humeral head in the plane shown. The maximum output torque $\tau_{33}$ during joint rotation is related to the joint position, including the flexion/extension angle $\theta_{31}$, the abduction/adduction angle $\theta_{32}$, and the rotation angle $\theta_{33}$. Of these, $\theta_{32}$ and $\theta_{33}$ have a greater effect on $\tau_{33}$. The relationship between $\tau_{33}$ and $\theta_{32}$, and $\theta_{33}$ is discussed, denoted as $\tau_{33}=f(\theta_{32},\theta_{33})$.

As shown in Fig. \ref{fig7.10}(g), the glenohumeral joint abducts at an angle and the humerus is rotated from $ON$ to $ON'$. The tendons of the subscapularis and infraspinatus will slide over the humeral head and no longer pass through the centre $O$ of the humeral head. The tendon insertion points $T_s$ and $T_i$ will move upwards with joint abduction. The moment arm driving the joint is $l_{m0}$, it is the projection of the radius $l_{12}$ of the truncated circle on a plane perpendicular to $ON'$. Point $M$ is the centre of the truncated circle, where the truncated circle pass through the contact points $M_{i}$, $M_{s}$ between the tendon and the humeral head. Point $R$ is on the truncated circle. 

Fig. \ref{fig7.10}(h) shows the section view of the joint on the symmetrical plane. The subscapularis tendon is attached from the motor $A$ to the point $T$ of the humeral head (projection of the tendon insertion point $T_s$ on the symmetrical plane). $AT$ passes through point $M$. As the glenohumeral joint abducts at an angle $\theta_{32}$, $OT_0$ rotates to $OT$, forming an angle $\theta_{32}$. The length of $TP$, $l_{15}$ is:
\begin{equation}
l_{15}=l_{14}sin\theta_{32}
\label{eq7.7}
\end{equation}

Where, $l_{14}$ is the length of the $OT$, it is known.
As $\Delta TPA$ and $\Delta MOA$ are similar, the length $l_{16}$ of $MO$ can be calculated as:
\begin{equation}
l_{16}=\frac{l_{11}l_{15}}{l_{11}+l_{14}cos\theta_{32}}
\label{eq7.8}
\end{equation}

Where $l_{11}$ is the length of the $OA$, it is known.

$l_{12}$ can be calculated as:
\begin{equation}
l_{12}=\sqrt{l_{18}^2-l_{16}^2}
\label{eq7.9}
\end{equation}

Where $l_{18}$ is the radius of the humeral head, it is known.

In $\Delta AOT$, according to the cosine theorem, The length of $AT$, $l_{17}$ can be calculated:
\begin{equation}
l_{17}=\sqrt{l_{11}^2+l_{14}^2-2l_{11}l_{14}\cos(\pi-\theta_{32})}
\label{eq7.10}
\end{equation}

According to the sine theorem, $\angle ATO$, $\theta_n$ can be calculated as:
\begin{equation}
\theta_n=arcsin(\frac{l_{11}\sin(\pi-\theta_{32})}{l_{17}})
\label{eq7.11}
\end{equation}

Under the action of the subscapularis tendon, the joint is rotated at $\theta_{33}$ from the position shown, as in Fig. \ref{fig7.10}(i). The tendon will slide from the dashed position to the solid position. The projection $OT$ of $OT_s$ will be decreased to $OT_1$. The length of $OT_1$, $l'_{14}$ can be calculated as:

\begin{equation}
{l'}_{14}=\frac{l_{14}\cos(\theta_{r0}+\theta_{33})}{cos\theta_{r0}}
\label{eq7.13}
\end{equation}

Where, $\theta_{r0}$ is the initial angle between $T_sO$ and $TO$, it is known.

 The moment arm driving the joint, $l_{m0}$ can be calculated as:
\begin{equation}
l_{m0}=l_{12}cos\theta_n
\label{eq7.6}
\end{equation}

Combining equations \ref{eq7.9}, \ref{eq7.11} and \ref{eq7.6}, by replacing $l_{14}$ with $l'_{14}$ in \ref{eq7.13}, the moment arm of the tendon driving the joint rotation, $l_{m0}$, can be obtained. Thus, $\tau_{33}=F_{t6} l_{m0}=f(\theta_{32},\ \theta_{33})$ can be obtained.

Given the symmetrical arrangement of the infraspinatus and subscapularis muscles, the infraspinatus tendon's analysis can be approached similarly. Given $F_{t6}$ = 600 N ((Maxon EC 4-pole 22mm 90W motors with 1:128 gearboxes are used to drive the infraspinatus and subscapularis tendon), based on the prototype's parameters, the relationship between $\tau_{33}$ and $\theta_{32}$, $\theta_{33}$ is obtained and shown in Fig. \ref{fig7.11}(c). It can be observed that as $\theta_{32}$ approaches $90\degree$, $\tau_{33}$ will approach 0. This is because the insertion point of the tendons of the infraspinatus and subscapularis has reached its highest position, where the moment arm $l_{m0}$ driving the joint rotation is close to 0 and unable to output torque. In the biological arm, the rotation of the glenohumeral joint at this point is driven by other muscles, but the torque is reduced in this position.

Table \ref{tab7.1} documents the maximum joint torques for each specific movement of the glenohumeral joint. To prevent prototype damage, maximal torques for each motion were not tested. Rather, the prototype's glenohumeral joint performance was assessed via practical operational testing.

\begin{table}[htb]
\caption{Comparison of robotic arm's joint torque with humans (without wrist and hand).}
\footnotesize
\begin{center}
\begin{tabular}{l l l l l}
\toprule
Motion & Joint torque & Human arm & Percentage$^1$ & Power$^2$\\
\midrule
Flexion $\tau_{31f}$ & 35 Nm & 93.7 Nm & 37.4\% & 63 W\\
Extension $\tau_{31e}$ & 34.7 Nm & 76.9 Nm & 45.1\% & 47.3 W \\
Abduction $\tau_{32b}$ & 54 Nm & 92.4 Nm & 58.4\% & 141.5 W \\
Adduction $\tau_{32d}$ & 35 Nm & 48.5 Nm & 72.2\% & 47.3 W \\
Extension $\tau_{33}$ & 18 Nm & 40 Nm & 45\% & 47.3 W \\

\bottomrule
\end{tabular}
\begin{tablenotes}
     \small
      $^1$ The percentage of joint torque realized compared to that of human joints.      
      $^2$ The rated output power of the actuators applied.
    \end{tablenotes}
\label{tab7.1}
\end{center}
\end{table}

\section{Performance and Validation}

This section presents a validation of the motion and manipulation performance of the proposed robotic arm. It includes an evaluation of the active range of motion of each joint and an object manipulation test. Mechanical intelligence, such as the self-locking mechanism and multi-joint tendon interactions, will be validated via a simplified robotic arm model.

\subsection{Range of motion}

\begin{figure}[htb]
\centerline{\includegraphics[width=0.65\textwidth]{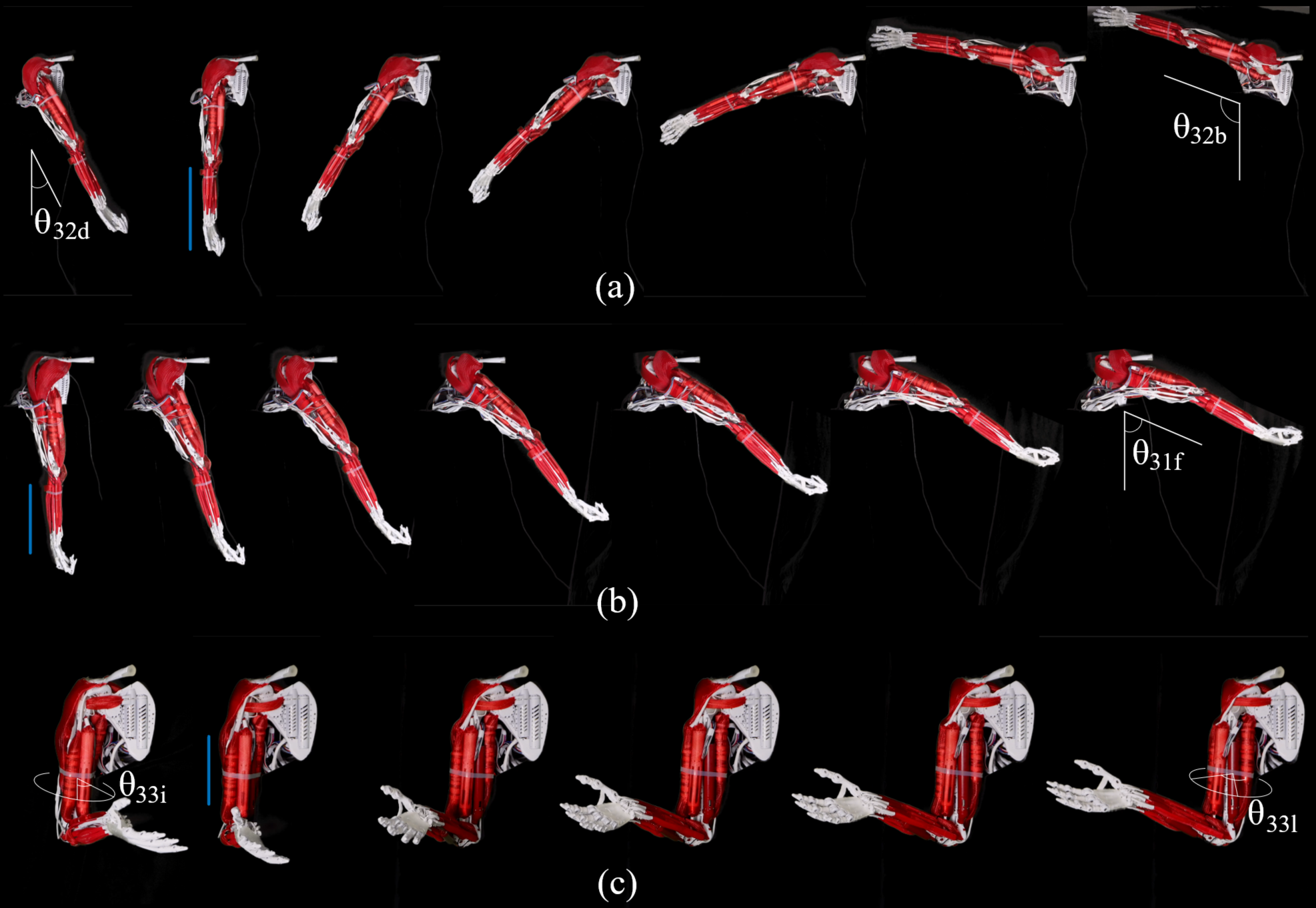}}
\caption{Range of motion test for each motion in the proposed robotic arm, glenohumeral joint adduction/abduction (a), flexion (b), rotation (c).}
\label{fig9.16}
\end{figure}

\begin{figure}[htb]
\centerline{\includegraphics[width=0.65\textwidth]{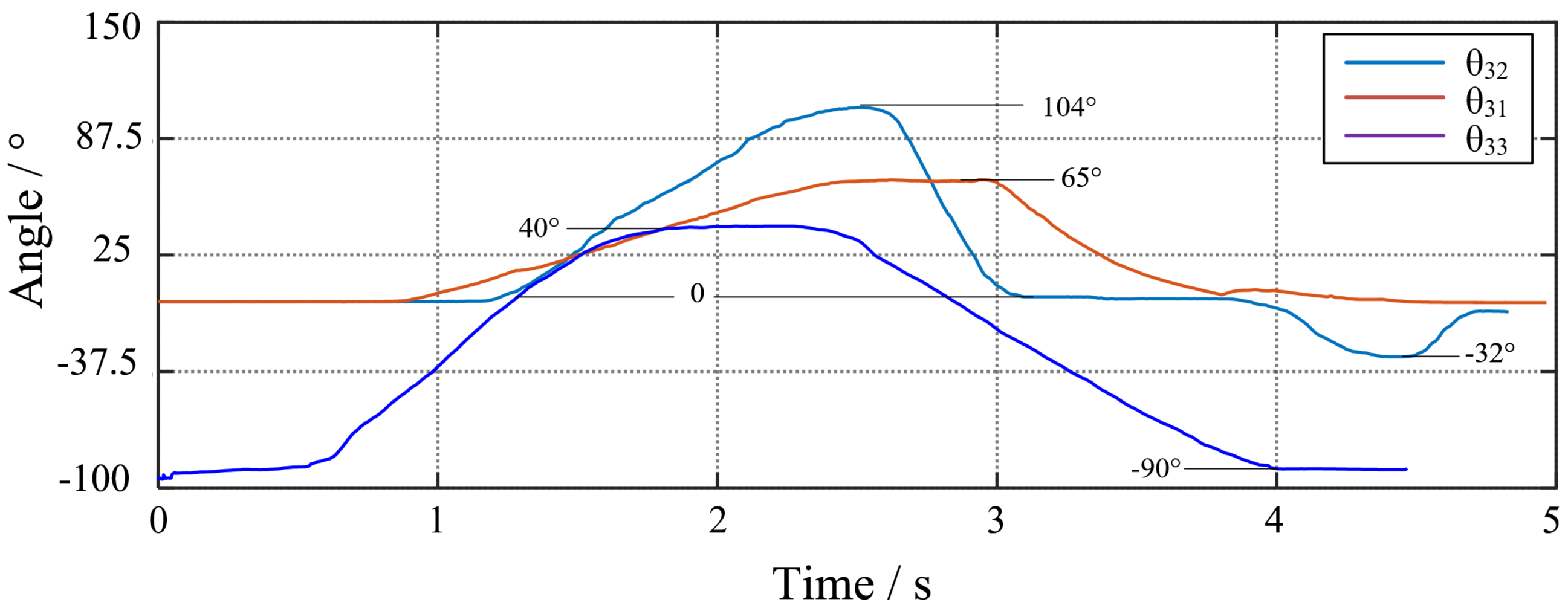}}
\caption{Data recorded of the range of motion test.}
\label{fig9.17}
\end{figure}

To record each active joint motion, the scapula of the robotic arm is fixed to the platform, and a gyroscope is used. Before each experiment, the gyroscope is calibrated, and its position is modified so that each measured joint motion corresponds to a change in the x-axis rotation angle of the gyroscope. The test results are shown in Fig. \ref{fig9.16} and Fig. \ref{fig9.17}, and the recorded ranges are compared with the data for the human arm, which is presented in Table. \ref{tab6.2}. Video 1 in the supplementary material presents the shoulder rotation motion test. Owing to the lack of musculature such as the pectoralis major (the torso is required) that facilitates glenohumeral joint flexion, the range of motion for the robotic arm is restricted.

\begin{table}[htb]
\centering
\caption{The joint range of motion of the robotic arm and human arm.}
\footnotesize
\begin{center}
\begin{tabular}{l l l}
\toprule
\multirow{2}*{Motion group of Glenohumeral joint} & \multicolumn{2}{c}{Joint range of motion} \\
             & robotic arm & human arm \\
\midrule
Extension (-) / Flexion (+) & -40-65\degree* & -60\degree-167\degree\\
Adduction (-) / Abduction (+) & -32\degree-104\degree & -29\degree-100\degree \\
Internal (-) / External (+) rotation & -90\degree-40\degree & -97\degree-34\degree  \\
\bottomrule
\end{tabular}
\begin{tablenotes}
      \small
      \centering
      *The range of motion for shoulder flexion and extension is inherently limited due to the lack of a torso, specifically the absence of chest and back musculature.
   \end{tablenotes}
\label{tab6.2}
\end{center}
\end{table}

\begin{figure*}[htb]
\centerline{\includegraphics[width=1\textwidth]{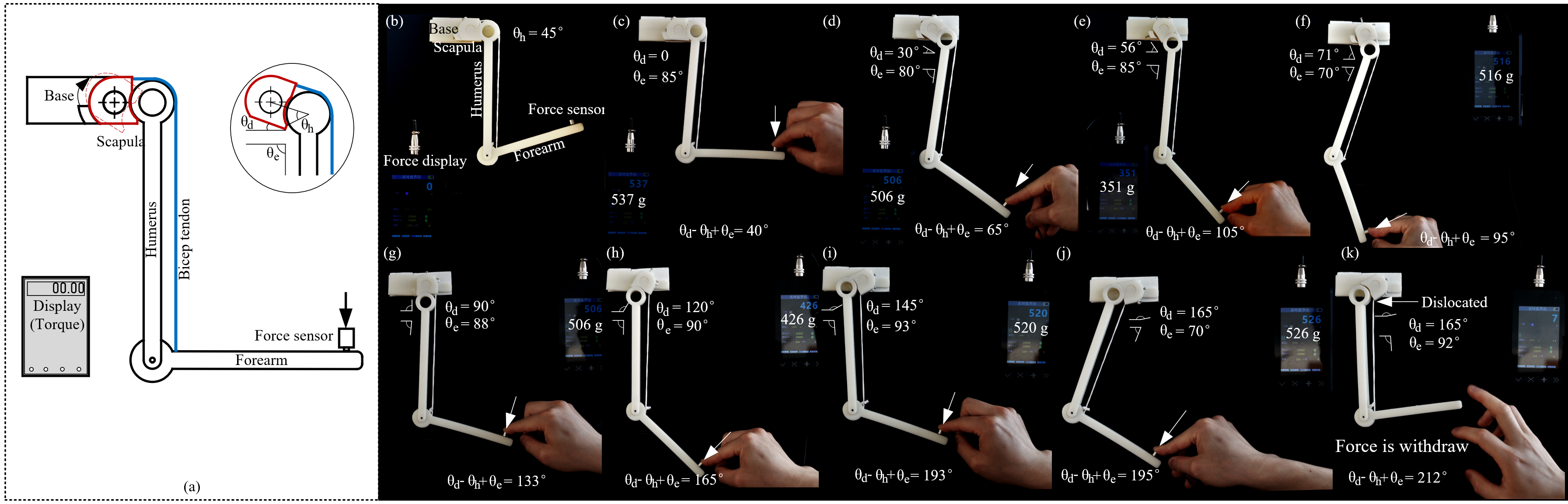}}
\caption{
Evaluation of tendon traversing multiple joints: (a) Schematic diagram of the experimental set-up; (b) Detailed depiction of the experimental configuration. Experimental outcomes for varying $\theta_a$: 0(c), 30\degree(d), 56\degree(e), 71\degree(f), 90\degree(g), 120\degree(h), 145\degree(i), 165\degree(j). (k) Observational results upon reaching the mechanism's failure condition.}
\label{fig8.18}
\end{figure*}

\subsection{Load capacity}

\begin{figure}[htb]
\centerline{\includegraphics[width=0.7\textwidth]{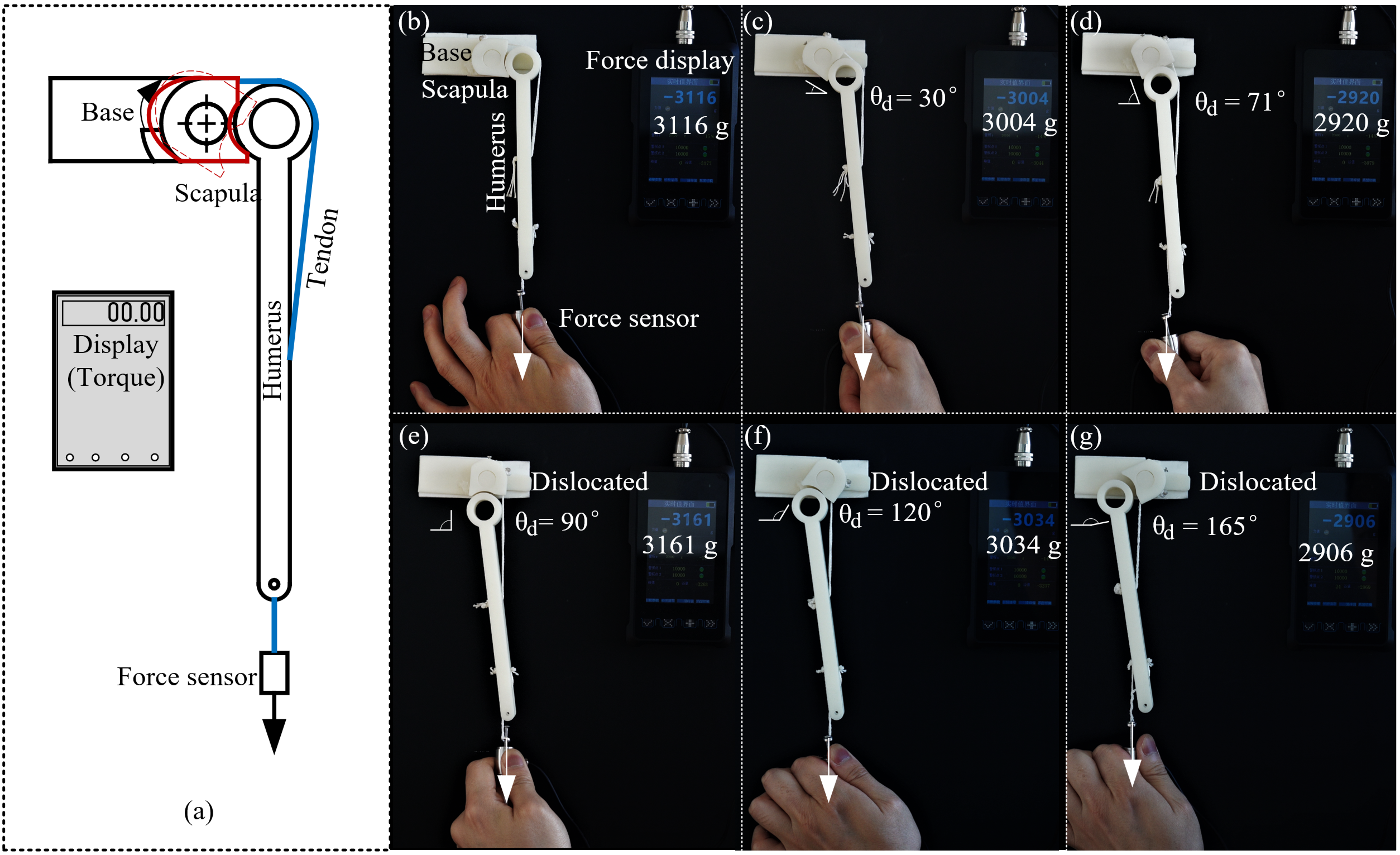}}
\caption{The experiment of the self-lock mechanism: (a) Schematic diagram of the experimental set-up. Experimental outcomes for varying $\theta_a$: 0(b), 30\degree(c), 71\degree(d), 90\degree(e), 120\degree(f), 165\degree(g).}
\label{fig8.19}
\end{figure}

\begin{figure}[htb]
\centerline{\includegraphics[width=0.5\textwidth]{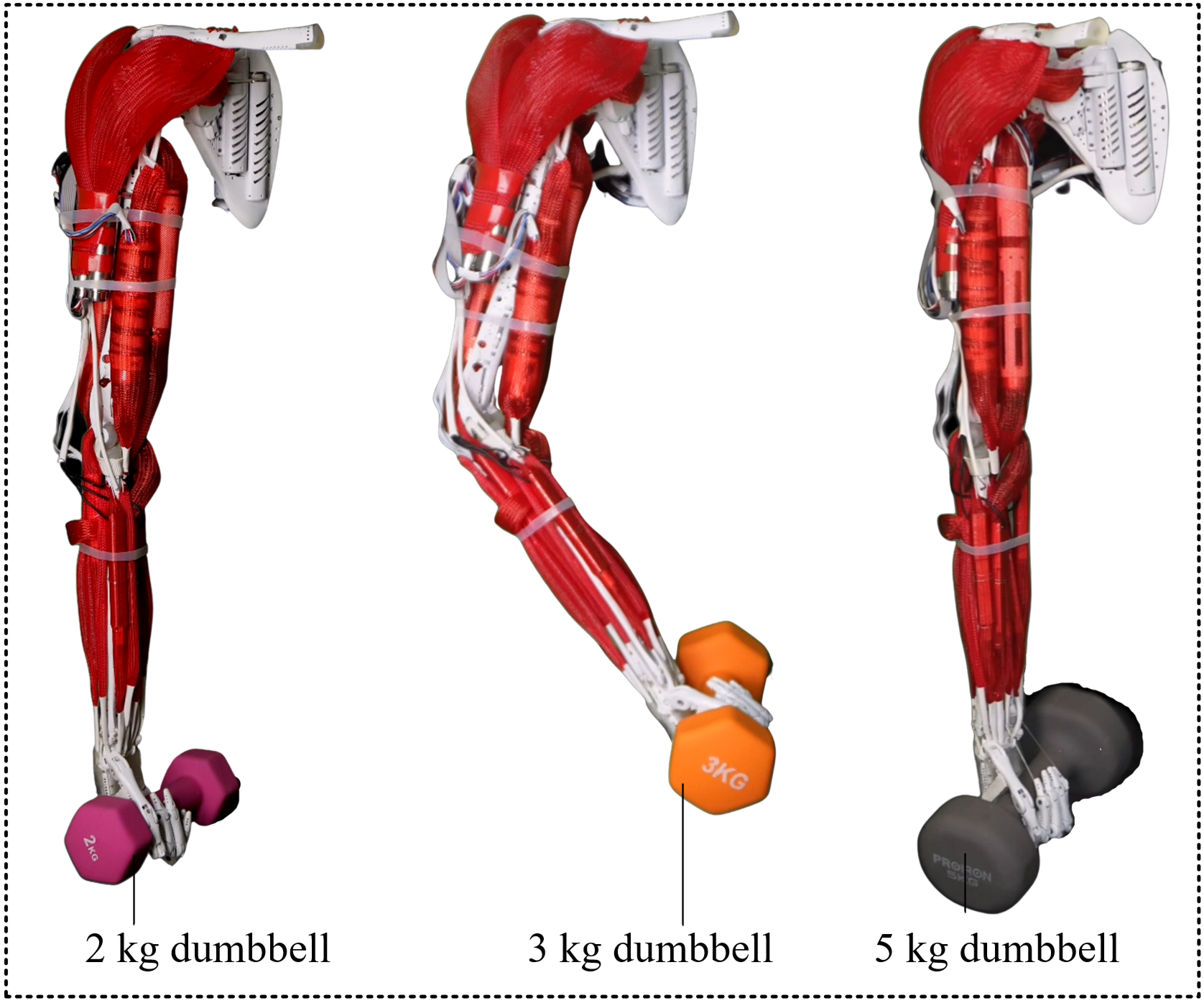}}
\caption{The completed robotic arm prototype holds the dumbbells (2kg, 3kg,5kg).}
\label{fig8.17}
\end{figure}

Firstly, the biceps muscle crosses both the elbow joint and the glenohumeral joint, whether this mechanical intelligence principle can improve the stability of the glenohumeral joint will be validated. The experimental apparatus, shown in Fig. \ref{fig8.18}(a), incorporates a fixed base on which the scapula is hinged, allowing adjustable rotation and angular constraint. A simplified arm model with the humerus and forearm hinged at the elbow joint was formed, retaining only the long head of the biceps tendon originating from the scapula and inserted into the forearm. Adjustments can be made to the scapula's angle ($\theta_{d}$) and the tendon's length during the experiment. Fig. \ref{fig8.18}(b) illustrates the experimental design.

A force sensor applied a perpendicular force to the distal forearm during the experiment. The test was conducted iteratively, modifying $\theta_{d}$ and adapting the tendon length accordingly. Fig. \ref{fig8.18}(c) to (i) demonstrate instances where $\theta_{d}$ equals 0, 30°, 56°, 71°, 90°, and 120° respectively. These scenarios did not meet the critical failure condition of the equilibrium system, i.e., $\theta_{d}-\theta_{s}+\theta_{e}<180\degree$.

Fig. \ref{fig8.18}(j) presents a case where $\theta_{d}=165\degree$ and $\theta_{d}-\theta_{h}+\theta_{e}=195\degree>180\degree$. Nevertheless, due to considerable friction between the humeral head and the scapula (omitted in theoretical calculations), the system maintained marginal stability when force was applied to the distal forearm. When the force was removed, the reduction in positive pressure caused friction between the humeral head and the scapula to approach zero, rendering the system unstable. Selected experimental results are presented in Video 1 within the supplementary material. The resultant dislocation of the glenohumeral joint occurred at $\theta_{d}-\theta_{h}+\theta_{e}=212\degree>180\degree$.

All tendons enveloping and spanning the glenohumeral joint (such as the Deltoid, Subscapularis, Biceps, etc.) are capable of integrating the joint's ball-and-socket structure to attain self-locking mechanical intelligence. As long as the structural failure's critical conditions are unmet, the joint undergoing a vertical downward force allows the tendon to press the humeral head into the glenoid. An escalation in the external force heightens the force compressing the humeral head, hence rendering the glenohumeral joint more stable.

To substantiate this property, Fig. \ref{fig8.19}(a) shows a simplified model of the robotic glenohumeral joint. An identical experimental setup was used, as previously employed, validates that multi-joint tendons can augment joint stability. The representation of a tendon enveloping the glenohumeral joint involves using a tendon that originates from the scapula and is inserted into the humerus (shown in blue). This experiment involves the omission of the forearm, and attaching a force sensor to the distal humerus with a cable. This allows the manual application of a vertical downward tension to the humerus, with the force's magnitude being exhibited on the display.

The alteration of the scapula's angle, i.e., $\theta_d$, entails a concurrent adjustment of the tendon's length to ensure its tightness when the humerus is pulled vertically downward. Under different $\theta_d$, force was applied to the distal humerus to observe joint dislocation. The experimental results at $\theta_d$ of 0, 30°, 71°, 90°, 120°, 165° are shown in Fig. \ref{fig8.19}(a)-(g), respectively.

The observation reveals that the self-locking structure is effective and the joint remains stable when $\theta_d<$90\degree. However, when $\theta_d>$90\degree, the tendon fails to apply the force necessary to push the humeral head into the glenoid under the external force. As the tendon stretches, the self-locking structure fails, resulting in joint dislocation. This condition worsens when $\theta_d$ continues to increase, culminating in a complete dislocation of the joint when $\theta_d$ equals 165\degree.

Further to demonstrate the load capabilities of the biomimetic robotic glenohumeral joint, a non-destructive experiment was conducted. The test involved lifting various weights using the fully assembled arm prototype. As shown in Fig. \ref{fig9.17}, the robotic arm successfully lifted three different weights, specifically 2kg, 3kg, and 5kg dumbbells. Notably, no dislocations were observed during the lifting process.

\subsection{Manipulative experiments in restricted environments}

\begin{figure*}[htb]
\centerline{\includegraphics[width=1\textwidth]{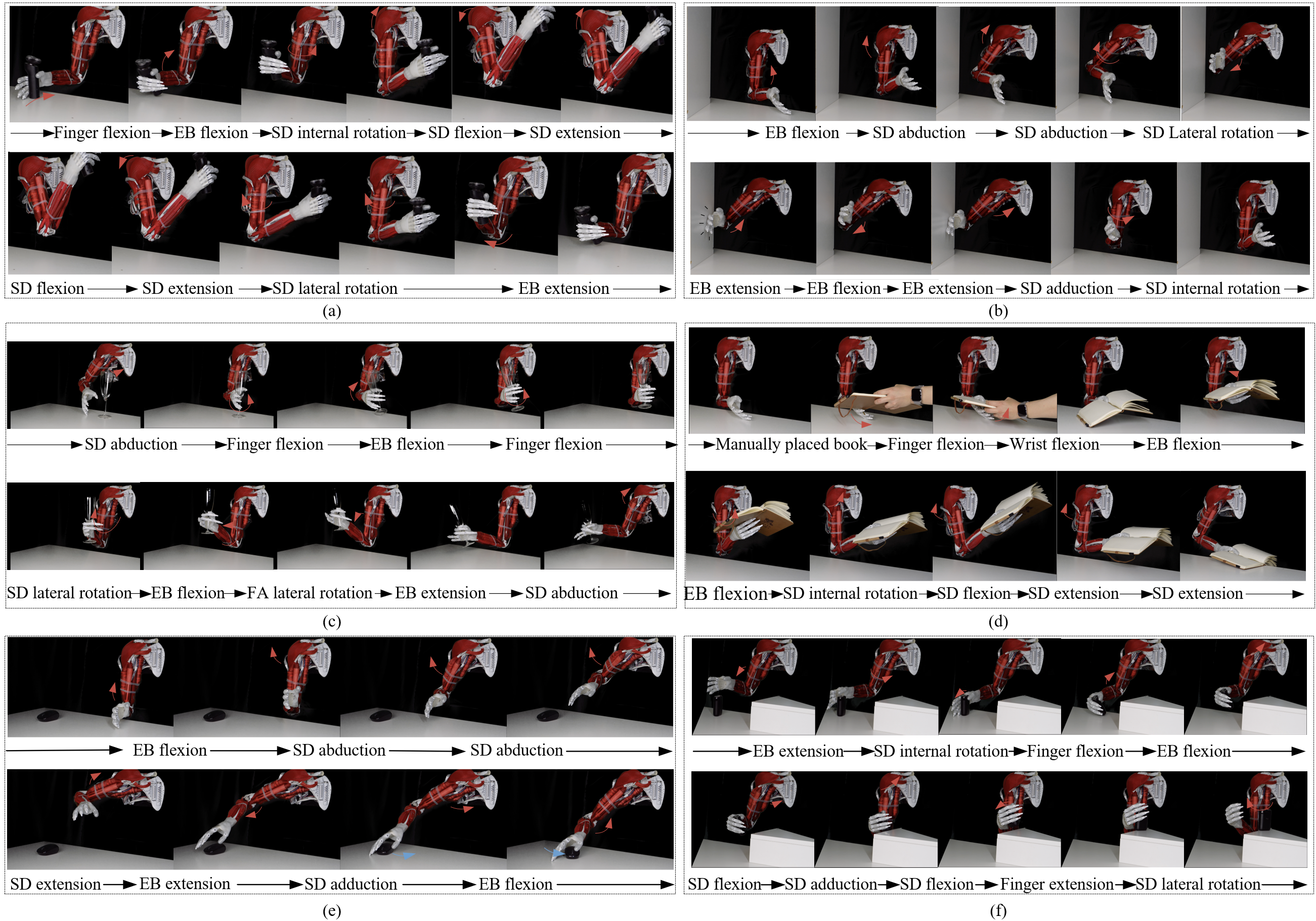}}
\caption{Manipulative Experiments (SD denotes Shoulder; EB denotes Elbow; FA denotes Forearm): (a) Shaving Imitation Test; (b) Door Knocking Test; (d) Mouse Movement on Desk Test; (e) Item Transference to Platform Test. }
\label{fig8.4}
\end{figure*}

\subsubsection{\textbf{Shaving Simulation}}
Fig. \ref{fig8.4}(a) portrays the robotic arm prototype replicating the human action of picking up a razor and executing a reciprocal shaving motion. Upon grasping the razor, the robotic arm elevates it to a ‘facial' proximity using shoulder movements. The back-and-forth motion approximates the act of shaving. Eventually, the razor is returned to the table. This task's complexity lies in the necessity for the robotic arm to perform a large shoulder internal rotation to correctly position the razor beneath the 'face'. The corresponding experimental footage can be found in Video 3 of the supplementary material.

\subsubsection{\textbf{Simulating Door Knocking}}
Fig. \ref{fig8.4}(b) features a wooden board positioned adjacent to the robotic arm to mimic a door. Emulating the joint movements during a human hand's door knock, the robotic arm performs the depicted motion sequence. The process starts with positioning the back of the hand close to the ‘door', swiftly executing a knock through elbow flexion and extension and then returning to its initial position. The test's challenge resides in correctly positioning the back of the hand within a confined space, specifically the central region of the ‘door', for maximal acoustic impact. Inaccurate hand positioning, too close to the ground, for instance, can result in the back of the hand striking the floor and causing test failure. Additionally, the robotic arm must avoid ground contact during hand positioning. With less than 25 cm separating the 'door' and the robotic arm, successful hand positioning within the correct area requires considerable compactness from the robotic arm. The corresponding experimental footage can be found in Video 4 of the supplementary material.

\subsubsection{\textbf{Goblet Lifting and Clinking Simulation}}
Fig. \ref{fig8.4}(c) depicts the robotic arm prototype imitating the action of a human arm lifting a goblet and performing a toasting motion. Initially placed on a table, the goblet is grasped by the flexing fingers and thumb of the robotic hand. Subsequent elbow flexion lifts the goblet, followed by forearm rotation simulating the act of clinking the goblet in a toast. The task's complexity lies in the significant degree of shoulder internal rotation and the need for a firm grip on the goblet to prevent slipping during actions such as lifting and tilting. The corresponding experimental footage can be found in Video 5 of the supplementary material.

\subsubsection{\textbf{Book Handling}}
Fig. \ref{fig8.4}(d) (video 6 in the supplementary material) shows the robotic arm receiving a book manually, which it secures by flexing its thumb. The book is lifted via wrist flexion and positioned in a reading-like posture. Subsequently, the robotic arm gently returns the book to the desk.

\subsubsection{\textbf{Mouse Operation}}
Fig. \ref{fig8.4}(e) (video 7 in the supplementary material) presents a scenario where a mouse is placed on a desk in front of the robotic arm, near the edge. The robotic arm positions its hand over the mouse, mimicking the joint movements of a human arm operating a mouse. Then, the glenohumeral joint adducts to grip the mouse, with the robotic arm effectively manoeuvring the mouse through elbow flexion. The task's challenge lies in the mouse's initial positioning, located on the desk edge and distally on the arm's right side. This arrangement emulates a real-world situation, with the arm at the edge of the desk, requiring abduction and shoulder joint extension to reach the mouse, pushing the shoulder joint close to its limits.

\subsubsection{\textbf{Object Transference to a Platform}}
Fig. \ref{fig8.4}(f) (video 8 in the supplementary material) illustrates the robotic arm conveying an object onto a platform of varying heights. The process commences with the robotic arm gripping the object using its fingers and thumb. Through shoulder flexion, the object is repositioned onto the platform, followed by its release from the robotic hand. The sequence concludes with the shoulder executing a lateral rotation. It is noteworthy that both the platform and the objects to be grasped are in immediate proximity to the robotic arm. The objects are situated near the table's edge, and the platform is positioned directly in front of the robotic arm without significant intervening distance. To circumvent contact with the platform and table, the robotic arm has limited operational space. The figures illustrate that the robotic arm retains a minimal spatial footprint throughout the test. When depositing items onto the platform, the forearm closely approaches the platform, potentially impeding proper object placement. This sequence highlights the robotic arm's capability to function within confined spaces and underscores its compact design. This benefit can be ascribed to the glenohumeral joint, which offers three degrees of freedom within a single joint.

\begin{figure*}[htb]
\centerline{\includegraphics[width=1\textwidth]{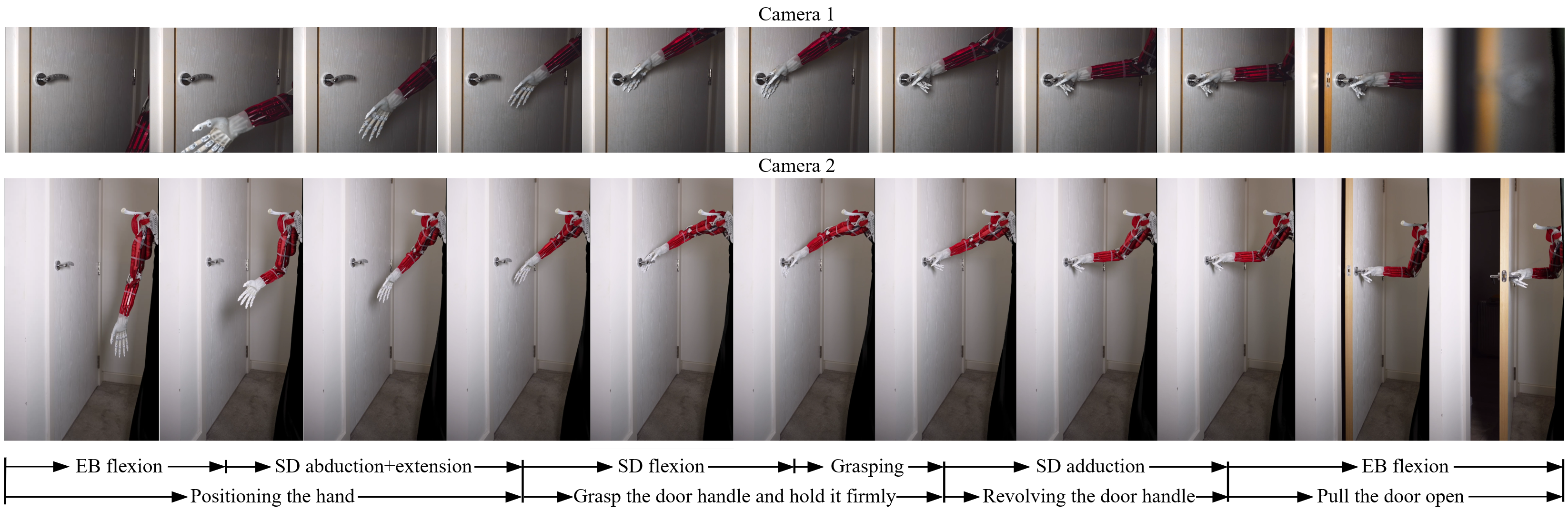}}
\caption{Test of opening the door (SD: shoulder; EB: elbow).}
\label{fig8.12}
\end{figure*}

\subsubsection{\textbf{Opening the door}}
Fig. \ref{fig8.12} demonstrates the robotic arm undertaking a door-opening task. Within the test. Initially, following a composite action sequence of the shoulder joint, the robotic hand is positioned on the door handle and subsequently secures a firm grip. Subsequently, the long head of the triceps instigates shoulder adduction to unlock the door by rotating the handle. Thereafter, the elbow flexes to facilitate door opening and extends for the closing action. In the experiment, a torque exceeding 1.5 Nm was required to unlock the door handle. The door-opening experiment serves as a widely used performance assessment for dexterous robots. While opening a door poses no challenge for an able-bodied individual, it may prove more difficult for someone with arm impairments. In conducting this test, the robotic arm faces the challenge of generating sufficient torque to rotate the door handle. When the human arm turns a door handle, downward pressure can be naturally applied to the handle by leveraging the body weight. However, the robotic arm prototype lacks a torso, preventing it from simply grabbing and pressing the handle downward. Instead, the robotic arm must securely grasp the handle before rotating it. Additionally, positioning the robotic hand in the appropriate location is another challenge, which requires the coordination of multiple motion sequences. Pulling the door also requires sufficient joint torque. This test serves to verify the robotic arm's proficiency in handling more demanding daily tasks. The corresponding experimental footage can be found in Video 9 of the supplementary material.

\section{Discussion}

\begin{table*}[htbp]
\caption{{Comparison of different robotic arms}}
\footnotesize
\begin{center}
\begin{tabular}{l l l l l l l l}
\toprule
 Name & Weight (kg)  & Payload (kg)  & Range of motion (\degree)$^A$ & Year & Driven method & Bio$^B$ \\
\midrule


Asimo \cite{shigemi2018asimo} & / & 0.5 & / & 2000 & Direct drive & No \\

Hubo 2 \cite{park2007mechanical} & / & 2 & / & 2009 &  Direct drive & No \\

Morgan et al.\cite{quigley2011low} & 11.4 & 2 & / & 2011 & Tendon+Timing Belt & No \\

R1 robot \cite{sureshbabu2017parallel} & / & 1.5 & / & 2017 & Direct drive & No \\

ABB-YuMi \cite{zahavi2018abb} & 9.1 & 0.5 & / & 2017 & Direct drive &  No \\

LIMS\cite{kim2017anthropomorphic} & 5.5 & 2.9 & /& 2017 & Tendon+Timing Belt & No \\

Tsumaki et al.\cite{tsumaki20187} & 2.9 & 1.5 & -180-180, -90-45, -180-180 & 2018 & Tendon & No \\

Reachy robot\cite{mick2019reachy} & 1.67 & 0.5 & / & 2019 & Direct drive &  No\\

LWH\cite{yang2019lwh} & 3.5 & 0.3 & -180-50, -45-180, -50-70 & 2019 & Direct drive & No \\

AMBIDEX \cite{choi2020hybrid}& 2.63 & 3 & / & 2020  & Tendon & No \\


Li et al.\cite{li2020modularization} & 2.2 & 1.5 & -70-270, -15-195, -70-270 & 2021 & Tendon & No \\



Kengoro\cite{asano2017design}  & / & / & -125-5, 0-120, -35-90 & 2017 & Tendon & Yes \\

Kenshiro\cite{potkonjak2011puller}  & / & / & -180-45, 0-122, -5-10 & 2019 & Tendon & Yes \\

\textbf{Proposed deisgn} & 4$^{C}$ & 4 & -40-65$^D$, -32-104, -40-90 & 2023 & Tendon & Yes \\ 

\bottomrule
\end{tabular}

 \begin{tablenotes}
        \footnotesize
        \item[]  $^A$Range of motion for glenohumeral joint extension(-)/flexion(+) and abduction(-)/adduction(+), lateral(-)/internal(+) rotation. $^B$Whether highly biomimetic robotics with biological joints or musculoskeletal designs. $^C$Weight of the proposed robotic arm including the forearm and hand and all the actuators (power supply and motor controller not included). $^D$The range of motion for shoulder flexion and extension is inherently limited due to the lack of a torso, specifically the absence of chest and back musculature.
      \end{tablenotes}
\label{tab4}
\end{center}
\end{table*}

The proposed glenohumeral joint design eschews the traditional design of a hinge joint with a rigid axis and draws on and replicates the biological structure of humans, including bones, ligaments, tendons and compliant actuators with biomuscular performance characteristics. The current design is in the early stages of development and there are still functional refinements to be made, but after a series of tests it is possible to identify several notable advantages over existing robotic arms:

Appearance: The design closely resembles the human glenohumeral joint. The inclusion of the deltoid muscle enables the robotic arm to closely mimic the human shoulder joint's aesthetic with realistic musculature, especially when clothed. This stands in contrast to conventional robotic arms that, even undergarments, often display an angular, non-anatomical shoulder structure, devoid of human muscle contours. In future plans, artificial skin will be added to the prototype to achieve a closer similarity in appearance and structure to the human arm. While this attribute does not necessarily augment performance, a human-like appearance is critical given the growing demand for domestic service robots, facilitating their seamless integration into familial settings.

Compactness: The bio-inspired glenohumeral joint in the proposed design offers three degrees of rotational freedom within a single compact joint, a striking divergence from traditional models that employ sequential rotational joints for the same range of motion. This streamlined design thereby elevates the capability of the entire robotic arm prototype to operate within limited spaces. The significance of dimensional constraints is apparent when the robotic arm functions in confined areas or close to objects, mirroring human tasks such as stir-frying at a stove or using a computer mouse. Overly long or large robotic limbs may resort to suboptimal and impractical postures under these conditions. Moreover, the proposed robotic arm employs a local tendon-driven approach, with all actuators mounted on the arm's main structure, mirroring human anatomy. This design offers enhanced compactness and fidelity to the human form, particularly when compared to remote-tendon-driven robots (including those utilizing pneumatic muscles), which necessitate the enclosure of actuators (or air pumps) within a device unrestricted by volume and mass considerations remotely.

Range of motion: Simultaneously achieving exceptional compactness, the proposed design replicates the range of motion closely akin to the human glenohumeral joint. 46.3\% flexion/extension (The absence of the muscles on the torso, such as the pectoralis major, results in this limitation. Subsequent modifications could address this flaw by integrating relevant torso muscles into the design), 105.4\% adduction/abduction and 99.2\% internal/external rotation were achieved respectively. The performance parameters of several existing traditional and high-fidelity robotic arms are outlined in Table. \ref{tab4}. A comparison reveals that compared to existing highly biomimetic robotic arms such as Kenshiro\cite{potkonjak2011puller}, the proposed robotic arm demonstrates a 766\% improvement in the shoulder's lateral/internal rotation, paralleling Kengoro\cite{asano2017design}. However, given the incorporation of scapular movement in Kengoro's shoulder, coupled with the lack of torso muscles in the proposed robotic arm, the range of shoulder flexion/extension is only 46.6\% of Kengoro's. As both the proposed robotic arm and Kengoro employ a design simulation based on the human skeletal-muscle system, there exists a possibility to extend the existing glenohumeral joint design and integrate the scapulothoracic joint\cite{sheikhzadeh2008three}, i.e., the joint between the scapula and torso. Such modifications in the proposed robotic arm, particularly the inclusion of scapular motion, could enhance its range of motion by a third \cite{aliaj2022kinematic}, achieving a range of motion similar to a human shoulder while retaining the same form and size as the human arm.

Safety during HRI: The system, hinged and fixed by soft tissues, resembles a biological joint's tension-compression system, exhibiting damping and flexibility when subjected to external forces. This feature greatly improves safety, as limited external forces can be absorbed by the soft tissues. In cases of excessive external force, the joint can dislocate and recover independently. For irreversible dislocations caused by extreme external forces, manual repairs can be performed without replacing any parts, similar to an orthopaedic doctor repairing a dislocated human joint.

Load capacity: Compared to existing highly biomimetic robotic arms, the proposed design optimizes the load capacity. While conventional robotic arms using hinge joints easily achieve load ability, biomimetic designs with biological joints, such as ECCE\cite{potkonjak2011puller} and Roboy robot\cite{trendel2018cardsflow}, can become unstable. By observing the demonstration video, the vibration and instability of the joint can be observed at the end of the movement. The inclusion of soft tissues and mechanical intelligence in this design achieves stability akin to hinge joints, resulting in an enhanced load-carrying capacity. 

Payload: Performance parameters of various existing conventional and bio-inspired robotic arms are listed in Table. \ref{tab4}. The completed proposed robotic arm, excluding the motor drive and power supply, weighs approximately 4 kg (including the arm structure and all muscles). This weight is comparable to robots of equivalent capacity, such as LIMS\cite{kim2017anthropomorphic} (5.5 kg), Tsumaki et al.\cite{tsumaki20187} (2.9 kg), and LWH\cite{yang2019lwh} (3.5 kg). Notably, despite these similarities in weight, the proposed robotic arm exhibits a higher payload capacity of 4 kg.

The list of videos for testing the proposed robotic glenohumeral joint and demonstrating the capabilities of the robotic arm is provided in Table \ref{tab12}. The video is accessible via the following link: \url{https://youtu.be/ZT8rIcApPVo}.

\begin{table}[htb]
\caption{Multimedia extensions}
\footnotesize
\begin{center}
\begin{tabular}{l l}
\toprule
No.   & Description  \\
\midrule
Video 1 & Coupling stability of the humeroradial and glenohumeral joints \\
Video 2 & Glenohumeral joint rotation \\
Video 3 & Shaving simulation\\
Video 4 & Simulating door knocking \\
Video 5 & Goblet lifting and clinking simulation \\
Video 6 & Book handling \\
Video 7 & Mouse operation \\
Video 8 & Object Transference to a Platform \\ 
Video 9 & Opening the door \\ 
\bottomrule
\end{tabular}

\begin{tablenotes}
\centering
        \footnotesize
        \item[]  *The video is accessible via the following link: \url{https://youtu.be/ZT8rIcApPVo}
      \end{tablenotes}
      
\label{tab12}
\end{center}
\end{table}

\section{Conclusion}

This paper grounded in an in-depth study of human anatomy, has unveiled the inherent mechanical intelligence within the human shoulder and elucidated potential performance enhancements this knowledge can offer in designing robotic arms. The aim is not merely to propose a new paradigm for highly biomimetic robot design, but also to further affirm the functions and advantages of human tissue in anatomical research. As a discipline, anatomy has spent centuries proposing and validating the structure and function of human tissue. This research pioneers an approach to highlight the function and superiority of various human anatomical structures through the construction of physical robotic prototypes, thereby bridging the chasm between anatomical knowledge and practical application.

The methodology employed does not blindly or simplistically replicate human structures. Instead, it seeks to discover, encapsulate, and validate the ingenuity inherent in human structures throughout the replication process. This evolution is a journey into new robotic design directions, where both successes and failures provide invaluable learning opportunities. One initial challenge faced was the arrangement of the seven ligaments in the glenohumeral joint. In the initial stages, the ligaments were tightly stretched to ensure joint stability. However, motion testing of the prototype revealed that the range of motion fell considerably short of the design goal due to ligament length constraints. To rectify this, ligaments were lengthened to enable an effective range of motion, which, in turn, caused the joint to dislocate even without external forces. Further anatomical exploration underscored the critical role of seemingly trivial structures, such as the negative pressure within the joint. An attempt was initially made to replicate the negative pressure between the internal labrum of the glenohumeral joint and the humeral head, but technical and material limitations necessitated an alternative approach: the use of a spring-loaded preloaded ligament system.

This paper's significant contribution lies in affirming the viability and success of robotic arms that precisely mirror the structure of the human arm, offering a progressive strategy to augment existing robotic arm designs. For instance, tendons traversing multiple joints can augment the load-bearing capacity of glenohumeral joints, the employment of incomplete ball-and-socket structures can enhance joint range of motion, while the utilization of various soft tissues can offset stability deficiency. The final prototype realised a payload exceeding 4 kg and a load capacity of well over 5 kg, achieving a range of motion closely equivalent to a human joint, albeit with a confined flexion range due to the lack of a torso. The compactness was also validated through operational experiments. These insights and experiences can serve as a crucial benchmark for future designers, inspiring the creation of subsequent generations of highly biomimetic robotic arms.

As for future plans, the intention is to build upon the current design for further refinement. This could include adjustments to the torso section to facilitate scapula motion and the introduction of the pectoralis major and dorsal muscles to achieve full glenohumeral joint flexion/extension.

\bibliographystyle{./IEEEtran} 
\bibliography{./IEEEabrv,./IEEEexample}

\end{document}